\documentclass[runningheads]{llncs}
\usepackage{graphicx}
\usepackage{amsfonts}
\usepackage{amsmath}
\usepackage{amssymb}
\usepackage{amstext}
\usepackage{bm}
\usepackage{bbm}
\usepackage{booktabs}
\usepackage{chngcntr}
\usepackage{color}
\usepackage{colortbl}
\usepackage{float}
\usepackage[T1]{fontenc}
\usepackage{graphicx}
\usepackage[utf8]{inputenc}
\usepackage{mathtools}
\usepackage[inline,shortlabels]{enumitem}
\usepackage{pgfplots}
\usepackage{tikz}
\usepackage{times}
\usepackage{xspace}

\usepackage[symbol]{footmisc}
\renewcommand{\thefootnote}{\arabic{footnote}}

\newcommand\numberthis{\addtocounter{equation}{1}\tag{\theequation}}

\newcommand{\RANSAC}{\textsc{ransac}\xspace}

\newcommand{\LORANSAC}{\textsc{lo-ransac}\xspace}

\newcommand{\etal}{et al.\ }

\def\grevlex{GRevLex\xspace}
\def\Fig{Fig.~}
\def\Figs{Figs.~}
\def\Tab{Table~}
\def\Sec{Section~}
\def\Secs{Sections~}

\pgfplotsset{compat=newest}
\usetikzlibrary{plotmarks}
\usetikzlibrary{external}
\usepgfplotslibrary{patchplots}

\makeatletter
\newenvironment{customlegend}[1][]
{%
    \begingroup
    \pgfplots@init@cleared@structures
    \pgfplotsset{#1}%
}
{
  \pgfplots@createlegend
    \endgroup
}

\def\addlegendimage{\pgfplots@addlegendimage}
\makeatother

\newlength\fwidth 

\newcommand{\ma}[1]{\ensuremath{\mathtt{#1}}\xspace}
\newcommand{\ve}[1][x]{\ensuremath{\mathbf{#1}}\xspace}

\newcommand{\T}{{\!\top}}

\newcommand{\xd}[1][]{\ensuremath{\tilde{x}_{#1}}\xspace}
\newcommand{\yd}[1][]{\ensuremath{\tilde{y}_{#1}}\xspace}

\newcommand{\xp}[1][]{\ensuremath{x^{\prime}_{#1}}\xspace}
\newcommand{\yp}[1][]{\ensuremath{y^{\prime}_{#1}}\xspace}

\newcommand{\vX}[1][]{\ensuremath{\ve[X]_{#1}}\xspace}

\newcommand{\vx}[1][]{\ensuremath{\ve[x]_{#1}}\xspace}
\newcommand{\vxp}[1][]{\ensuremath{\ve[x]^{\prime}_{#1}}\xspace}
\newcommand{\vxd}[1][]{\ensuremath{\ve[\tilde{x}]_{#1}}\xspace}

\newcommand{\vXs}[1]{\ensuremath{ \{ \, \vX[i] \, \}^{#1}_{i=1}} \xspace}
\newcommand{\vxds}[1]{\ensuremath{ \{\, \vxd[i] \, \}^{#1}_{i=1}} \xspace}

\newcommand{\mA}[1][]{\ensuremath{\ma{A}_{#1}}\xspace}

\newcommand{\mH}{\ensuremath{\ma{H}}\xspace}

\newcommand{\mHhat}{\ensuremath{\ma{\hat{H}}}\xspace}

\newcommand{\mP}{\ensuremath{\ma{P}}\xspace}

\newcommand{\vl}{\ensuremath{\ve[l]}\xspace}
\newcommand{\vlinf}{\ensuremath{\ve[l]_{\infty}}\xspace}
\newcommand{\vu}{\ensuremath{\ve[u]}\xspace}
\newcommand{\vv}{\ensuremath{\ve[v]}\xspace}

\newcommand{\eg}{{\em e.g.}\xspace}

\newcommand{\rgn}[1][]{\ensuremath{\mathcal{R}_{#1}}\xspace}
\newcommand{\rgnd}[1][]{\ensuremath{\tilde{\mathcal{R}}_{#1}}\xspace}

\newcount\colveccount
\newcommand*\colvec[1]{
        \global\colveccount#1
        \begin{pmatrix}
        \colvecnext
}
\def\colvecnext#1{
        #1
        \global\advance\colveccount-1
        \ifnum\colveccount>0
                \\
                \expandafter\colvecnext
        \else
                \end{pmatrix}
        \fi
}

\newtoks\rowvectoks
\newcommand{\rowvec}[2]{%
  \rowvectoks={#2,}\count255=#1\relax
  \advance\count255 by -1
  \rowvecnexta}
\newcommand{\rowvecnexta}{%
  \ifnum\count255>0
    \expandafter\rowvecnextb
  \else
    \setlength\arraycolsep{1pt}     
    \begin{pmatrix}\the\rowvectoks\end{pmatrix}
  \fi}
\newcommand\rowvecnextb[1]{%
  \ifnum\count255>1     
    \rowvectoks=\expandafter{\the\rowvectoks&#1,}%
  \else
    \rowvectoks=\expandafter{\the\rowvectoks&#1}%
  \fi
    \advance\count255 by -1
    \rowvecnexta}

\definecolor{Gray}{gray}{0.9}
\newcolumntype{a}{>{\columncolor{Gray}}c}
\newcommand{\ra}[1]{\renewcommand{\arraystretch}{#1}}

\newcolumntype{L}[1]{>{\raggedright\let\newline\\\arraybackslash\hspace{0pt}}m{#1}}
\newcolumntype{C}[1]{>{\centering\let\newline\\\arraybackslash\hspace{0pt}}m{#1}}
\newcolumntype{g}[1]{>{\columncolor{Gray}\centering\let\newline\\\arraybackslash\hspace{0pt}}m{#1}}
\newcolumntype{R}[1]{>{\raggedleft\let\newline\\\arraybackslash\hspace{0pt}}m{#1}}

\pgfplotsset{
    legend image with text/.style={
        legend image code/.code={%
            \node[anchor=center] at (0.3cm,0cm) {#1};
        }
    },
}

\begin{document}
\pagestyle{headings}
\mainmatter 

\def\ACCV18SubNumber{364}  

\def\mytitle{Rectification from Radially-Distorted Scales}
\title{\mytitle} 

\titlerunning{\mytitle}
\authorrunning{J. Pritts \etal}

\author{James Pritts\inst{1,2} \and Zuzana Kukelova\inst{2} \and Viktor Larsson\inst{3}\textsuperscript{,*} \and Ond{\v r}ej Chum\inst{2}}

\institute{Czech Institute of Informatics, Robotics and Cybernetics (CIIRC), CTU in Prague \and Visual Recognition Group (VRG), FEE, CTU in Prague \and Department of Computer Science, ETH Zürich, Switzerland}

\maketitle

\begin{abstract}
This paper introduces the first minimal solvers that jointly estimate
lens distortion and affine rectification from repetitions of
rigidly-transformed coplanar local features. The proposed solvers
incorporate lens distortion into the camera model and extend accurate
rectification to wide-angle images that contain nearly any type of
coplanar repeated content. We demonstrate a principled approach to
generating stable minimal solvers by the Gr{\"o}bner basis method,
which is accomplished by sampling feasible monomial bases to maximize
numerical stability. Synthetic and real-image experiments confirm that
the solvers give accurate rectifications from noisy measurements if
used in a \RANSAC-based estimator. The proposed solvers demonstrate
superior robustness to noise compared to the state of the art. The
solvers work on scenes without straight lines and, in general, relax
strong assumptions about scene content made by the state of the
art. Accurate rectifications on imagery taken with narrow focal length
to fisheye lenses demonstrate the wide applicability of the proposed
method. The method is automatic, and the code is published
at \url{https://github.com/prittjam/repeats}.

%

\renewcommand{\thefootnote}{\fnsymbol{footnote}}
\footnotetext[1]{This work was done while Viktor Larsson was at Lund University.}
\renewcommand{\thefootnote}{\arabic{footnote}}
\end{abstract}

\section{Introduction}
This paper proposes minimal solvers that jointly estimate affine
rectification and lens distortion from local features extracted from
arbitrarily repeating coplanar texture. Wide-angle lenses with
significant radial lens distortion are common in consumer cameras like
the GoPro series of cameras. In the case of Internet imagery, the
camera and its metadata are often unavailable for use with off-line
calibration techniques.  The state of the art has several approaches
for rectifying (or partially calibrating) a distorted image, but these
methods make restrictive assumptions about scene content by
assuming, \eg, the presence of sets of parallel
lines \cite{Wildenauer-BMVC13,Antunes-CVPR17}. The proposed solvers
relax the need for special scene structure to unknown repeated
structures (see \Tab\ref{tab:state_of_the_art}). The solvers are
fast and robust to noisy feature detections, so they work well in
robust estimation frameworks like \RANSAC \cite{Fischler-CACM81}. The
proposed work is applicable for several important computer vision
tasks including symmetry detection \cite{Funk-ICCV17},
inpainting \cite{Lukac-ACMTG17}, and single-view 3D
reconstruction \cite{Wu-CVPR11}.

\begin{figure}[t!]
\begin{minipage}{0.25\columnwidth}
\parbox{\columnwidth}{
\centering
Input
}
\includegraphics[width=\columnwidth]{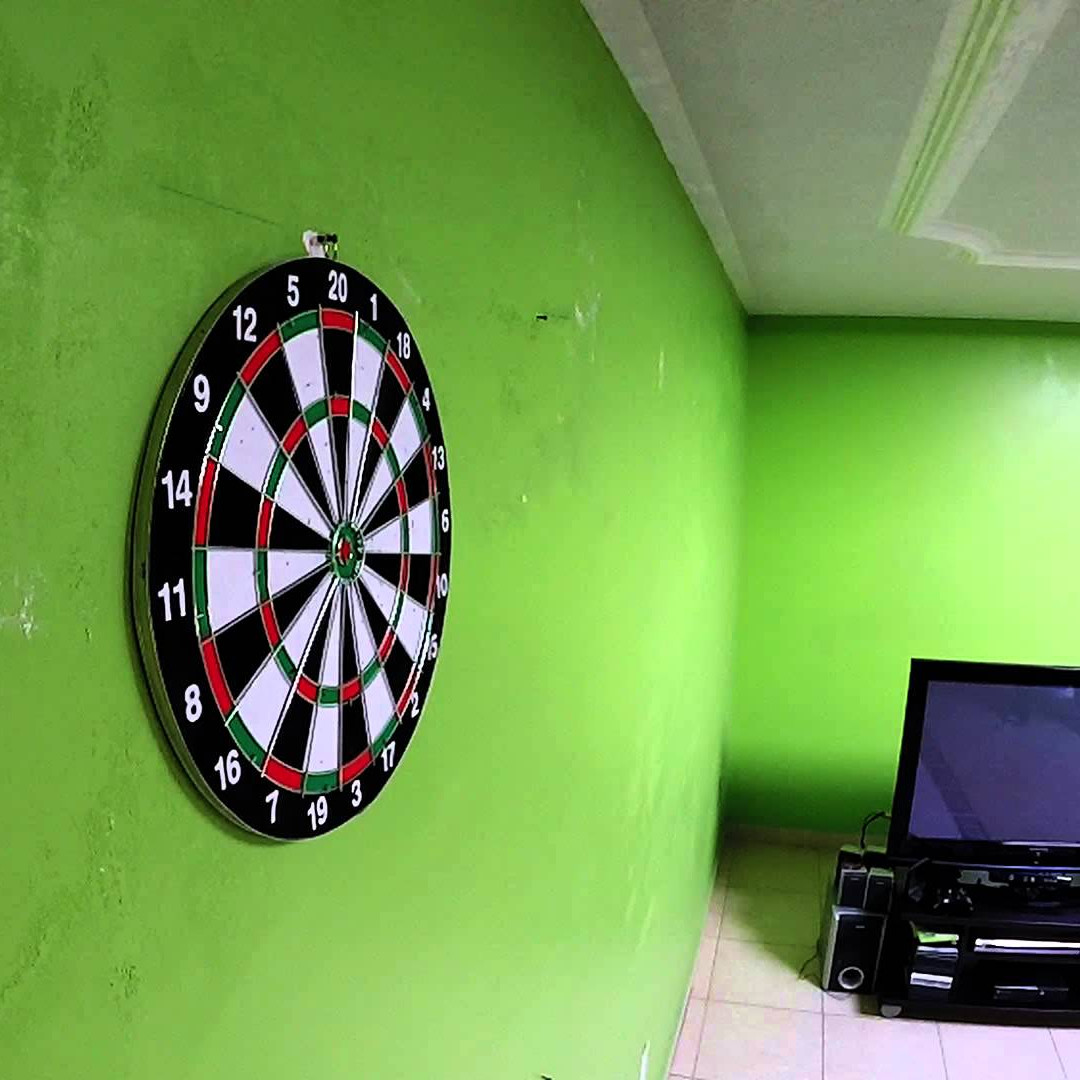}
\parbox{0.32\columnwidth}
{
\centering $222$
}
\parbox{0.32\columnwidth}
{
\centering $32$
}
\parbox{0.32\columnwidth}
{ \centering  $4$
}

\includegraphics[width=0.315\columnwidth]{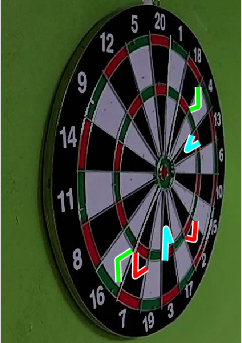}
\hfill
\includegraphics[width=0.315\columnwidth]{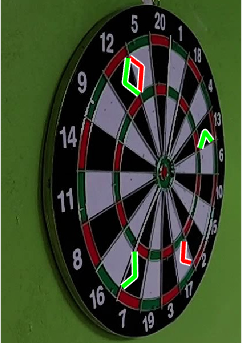}
\hfill
\includegraphics[width=0.315\columnwidth]{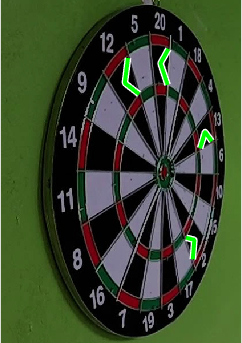}
\end{minipage}
\hfill
\begin{minipage}{0.72\columnwidth}
\parbox{0.32\columnwidth}
{
\centering $\mH_{222}\vl\lambda$
}
\parbox{0.32\columnwidth}
{
\centering $\mH_{32}\vl\lambda$
}
\parbox{0.32\columnwidth}
{
\centering $\mH_4 \vl\lambda$
}
\includegraphics[width=0.32\columnwidth]{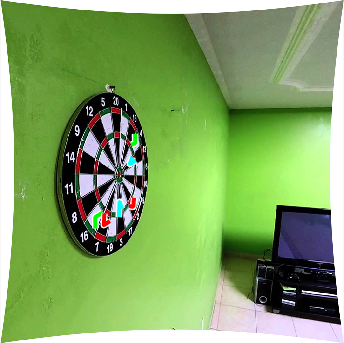}
\hfill
\includegraphics[width=0.32\columnwidth]{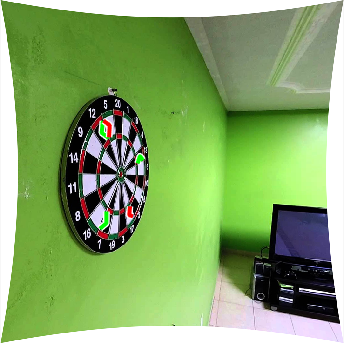} 
\hfill
\includegraphics[width=0.32\columnwidth]{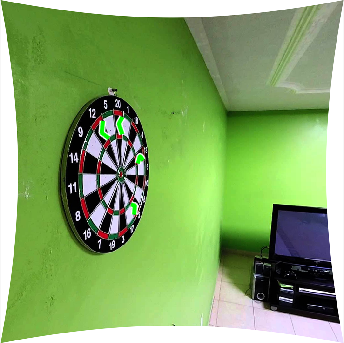}       

\includegraphics[width=0.32\columnwidth,trim={0 1.45cm 0 1.45cm},clip]{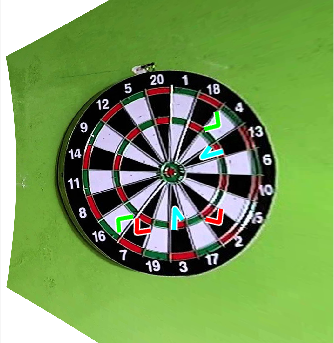}
\hfill
\includegraphics[width=0.32\columnwidth,trim={0 1.45cm 0 1.45cm},clip]{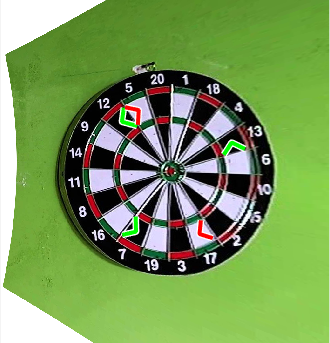}
\hfill
\includegraphics[width=0.32\columnwidth,trim={0 1.45cm 0 1.45cm},clip]{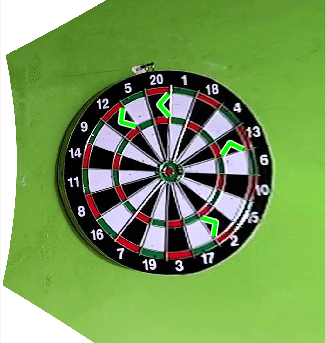}
\end{minipage}
\caption{\emph{Solver Variants.} (top-left image) The input to the method is
  a single image. (bottom-left triptych, contrast enhanced) The three
  configurations---$222,32,4$---of affine frames that are inputs to
  the proposed solvers variants. Corresponded frames have the same
  color. (top row, right) Undistorted outputs of the proposed solver
  variants. (bottom row, right) Cutouts of the dartboard rectified by
  the proposed solver variants. The affine frame
  configurations---$222,32,4$---are transformed to the undistorted and
  rectified images.}
\label{fig:solver_variants}
\end{figure}

The proposed solvers enforce the affine invariant that rectified
repeated regions have the same scale. We introduce three variants
(see \Fig{\ref{fig:solver_variants}}) that use different
configurations of coplanar repeated features as input, which allows
for flexible sampling during robust estimation.  Lens distortion is
parameterized by the division model, which
Fitzgibbon \cite{Fitzgibbon-CVPR01} first used to model lens
distortion and showed that it accurately models significant
distortions. The use of the division model is crucial because other
typical distortion models result in unsolvable constraint equations
(see Sec.~\ref{sec:radial_lens_distortion}). A fourth solver variant
is proposed that assumes the pinhole camera model. The pinhole variant
is also novel because it does not linearize the rectifying
transformation, which is the approach of the state of the
art \cite{Ohta-IJCAI81,Criminisi-BMVC00,Chum-ACCV10}.

The polynomial system of equations encoding the rectifying constraints
is solved using an algebraic method based on Gr{\"o}bner bases.
Automated solver generators using the Gr{\"o}bner basis
method~\cite{Kukelova-ECCV08,Larsson-CVPR17} have been used to
generate solvers for several camera geometry estimation
problems \cite{Kukelova-ECCV08,Larsson-CVPR17,Larsson-ICCV17,Kukelova-CVPR15,Pritts-CVPR18}. However,
straightforward application of automated solver generators to the
proposed constraints resulted in unstable solvers (see
Sec.~\ref{sec:experiments}). Recently,
Larsson \etal \cite{Larsson-CVPR18} sampled feasible monomial bases,
which can be used in the action-matrix
method. In \cite{Larsson-CVPR18} basis sampling was used to minimize
the size of the solver. We modified the objective
of \cite{Larsson-CVPR18} to maximize for solver stability. Stability
sampling generated significantly more numerically stable solvers (see
Fig.~\ref{fig:stability_and_1pxwarp}).

Several state-of-the-art methods can rectify from imaged coplanar
repeated texture, but these methods assume the pinhole camera
model \cite{Lukac-ACMTG17,Ohta-IJCAI81,Criminisi-BMVC00,Chum-ACCV10,Ahmad-IJCV18,Aiger-EG12,Zhang-IJCV12}. A subset of these methods use
algebraic constraints induced by the equal-scale invariant of
affine-rectified
space \cite{Ohta-IJCAI81,Criminisi-BMVC00,Chum-ACCV10}. These methods
linearize the rectifying transformation and use the Jacobian
determinant to induce local constraints on the imaged scene plane's
vanishing line. The Jacobian determinant measures the local
change-of-scale of the rectifying transformation. In contrast, the
proposed solvers are the first to directly encode the unknown scale of
a rectified region as the joint function of the measured region,
vanishing line, and undistortion parameter. The direct approach
eliminates the need for iterative refinement due to the linearization
of the rectifying homography.


\begin{table}[!t]
\centering
\ra{1}
\resizebox{\textwidth}{!}{
\begin{tabular}{@{} rC{20ex}C{18ex}C{16ex}g{16ex} @{} }
\toprule
& Wildenauer \etal \cite{Wildenauer-BMVC13} & Antunes \etal \cite{Antunes-CVPR17} & Pritts \etal \cite{Pritts-CVPR18} & Proposed \\
\midrule
Feature Type & fitted circles & fitted circles & affine-covariant & affine-covariant \\
Assumption & 3 \& 3 parallel lines & 3 \& 4 parallel lines & 2 trans. repeats & 4 repeats \\
Rectification & multi-model & multi-model & direct & direct \\
\bottomrule
\end{tabular}
}
\caption{\emph{Scene Assumptions.} Solvers \cite{Wildenauer-BMVC13,Antunes-CVPR17} require distinct sets of parallel scene lines as input and multi-model
estimation for rectification. Pritts \etal \cite{Pritts-CVPR18} is
restricted to scenes with translational symmetries. The proposed
solvers directly rectify from as few as 4 rigidly transformed repeats.}
\label{tab:state_of_the_art}
\end{table}

Pritts \etal \cite{Pritts-CVPR14} recover rectification with
distortion using a two-step approach: a rectification estimated from a
minimal sample using the pinhole assumption is refined by a nonlinear
program that incorporates lens distortion.  However, even with relaxed
thresholds, a robust estimator like \RANSAC \cite{Fischler-CACM81}
discards measurements around the boundary of the image since this
region is the most affected by radial distortion and cannot be
accurately modeled with a pinhole camera. Neglecting lens distortion
during the segmentation of good and bad measurements, as done during
the verification step of \RANSAC, can give fits that are biased to
barrel distortion
\cite{Kukelova-CVPR15}, which degrades rectification accuracy. Pritts
\etal \cite{Pritts-CVPR18} were the first to propose minimal solvers that 
jointly estimate affine rectification and lens distortion, but this
method is restricted to scene content with translational symmetries
(see
\Tab\ref{tab:state_of_the_art}). Furthermore, we show that the
conjugate translation solvers of \cite{Pritts-CVPR18} are more noise
sensitive than the proposed scale-based solvers (see
Figs.~\ref{fig:stability_and_1pxwarp}
and \ref{fig:ransac_sensitivity_study}).

There are two recent methods that affine-rectify lens distorted images
by enforcing the constraint that scene lines are imaged as circles
with the division model \cite{Wildenauer-BMVC13,Antunes-CVPR17}. The
input into these solvers are circles fitted to contours extracted from
the image.  Sets of circles whose preimages are coplanar parallel
lines are used to induce constraints on the division model parameter
and vanishing points.  These methods require two distinct sets of
imaged parallel lines (6 total lines for \cite{Wildenauer-BMVC13} and
7 for \cite{Antunes-CVPR17}) to estimate rectification, which is a
strong scene-content assumption. In addition, these methods must
perform a multi-model estimation to label distinct vanishing points as
pairwise consistent with vanishing lines.  In contrast, the proposed
solvers can undistort and rectify from as few as 4 coplanar repeated
local features (see \Tab\ref{tab:state_of_the_art}).

\section{Problem Formulation}
\begin{figure}[t!]
\includegraphics[width=0.145\columnwidth]{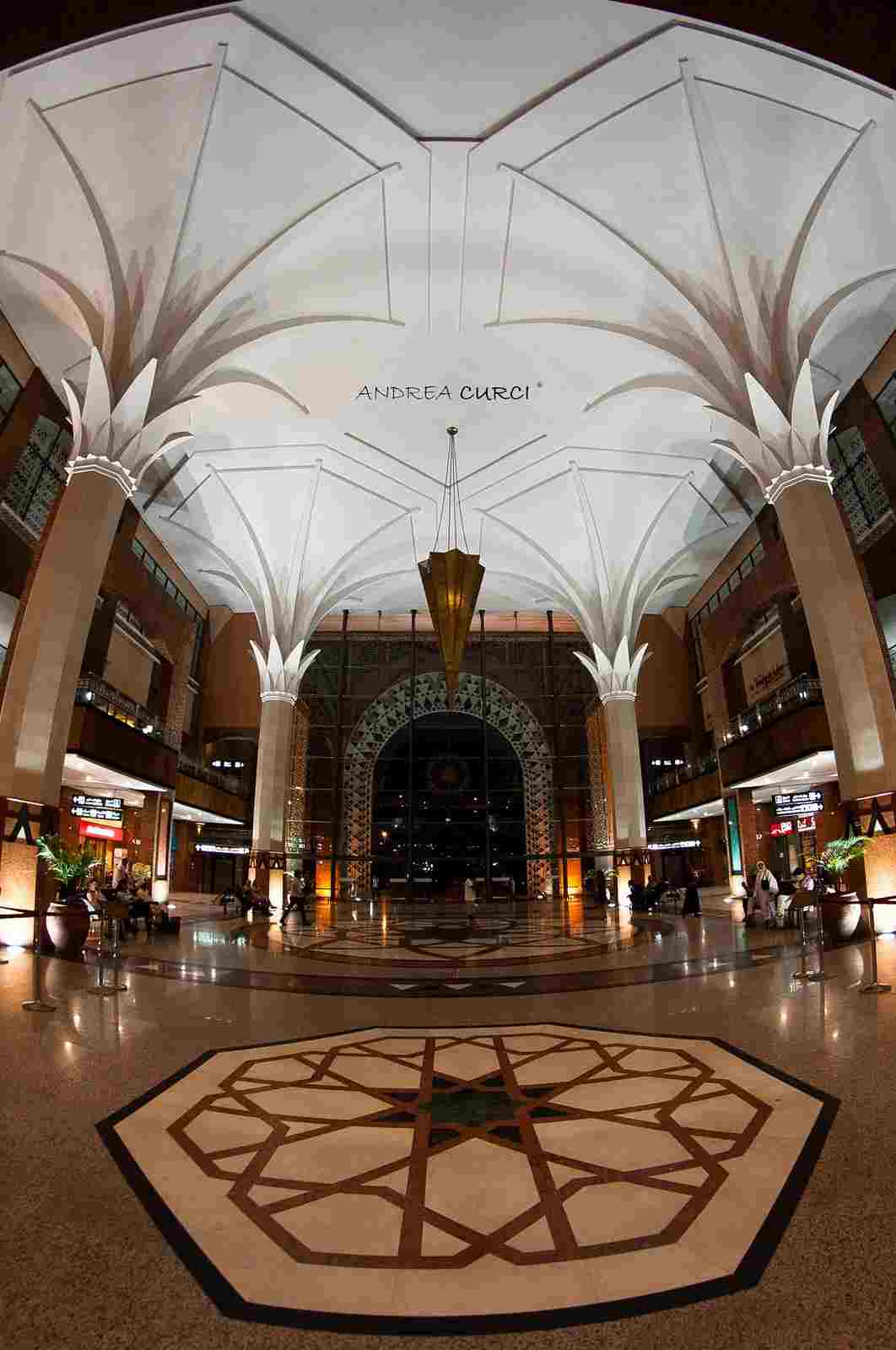}
\includegraphics[width=0.145\columnwidth]{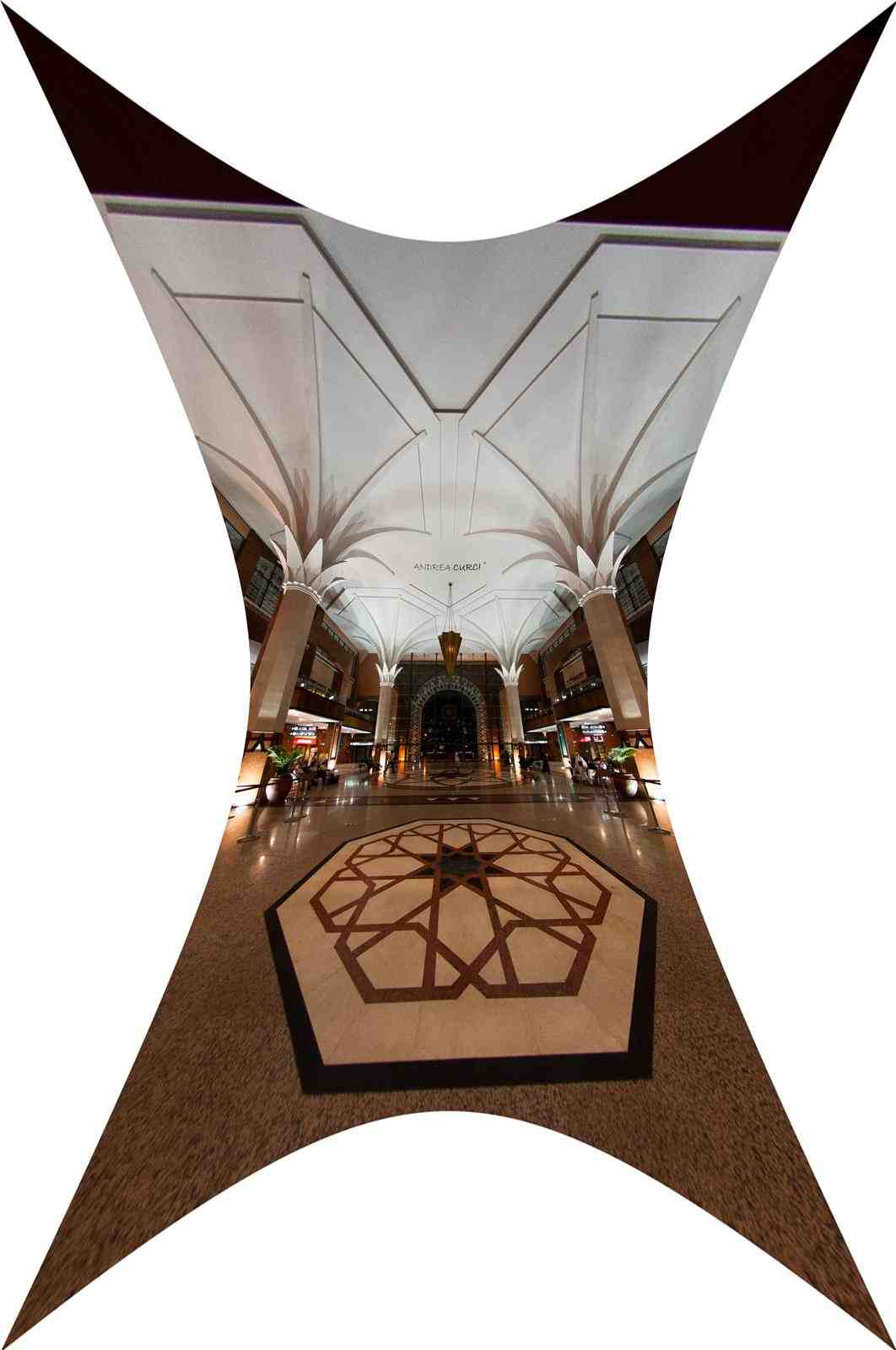}
\includegraphics[width=0.16\columnwidth]{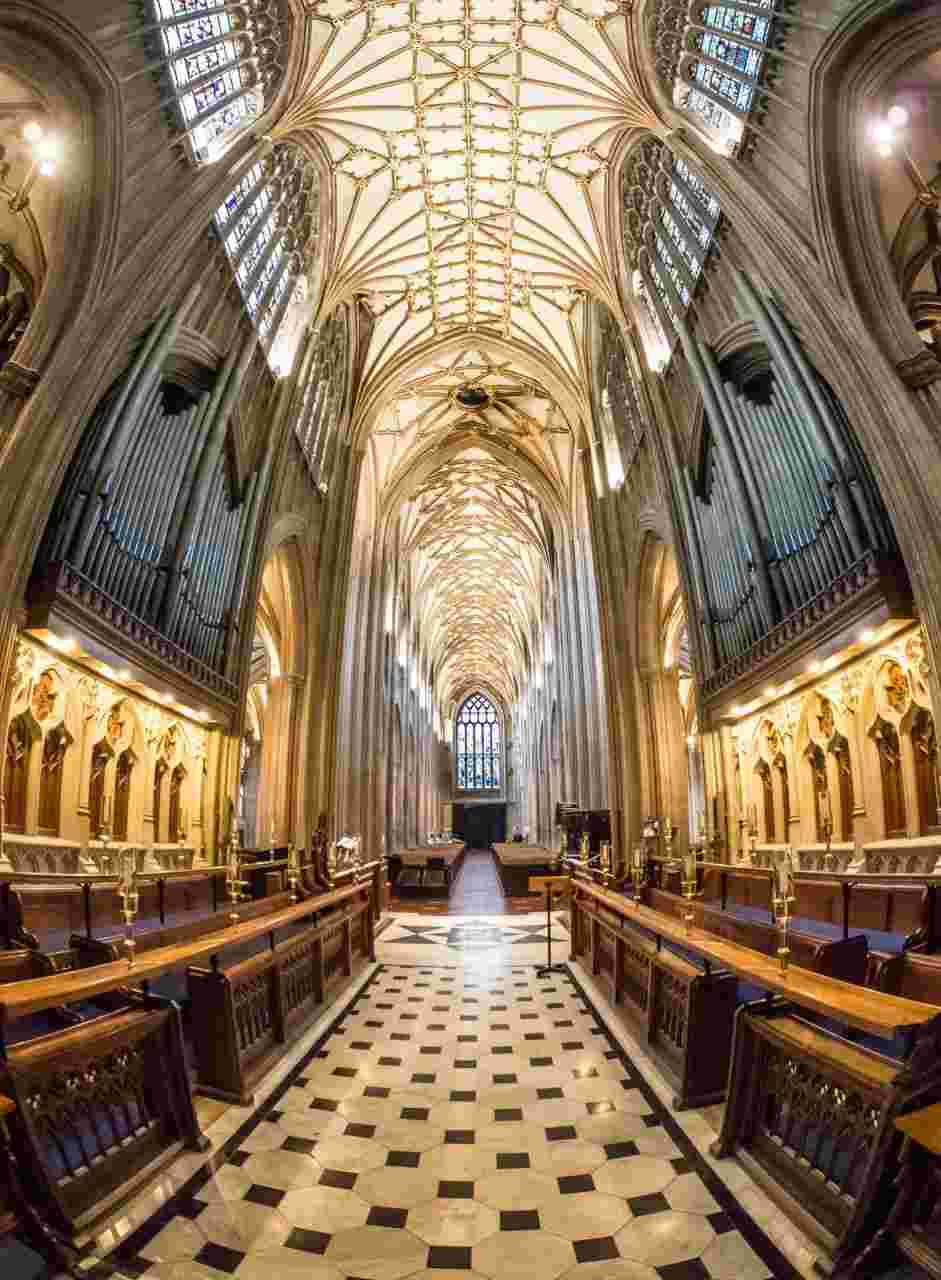}
\includegraphics[width=0.16\columnwidth]{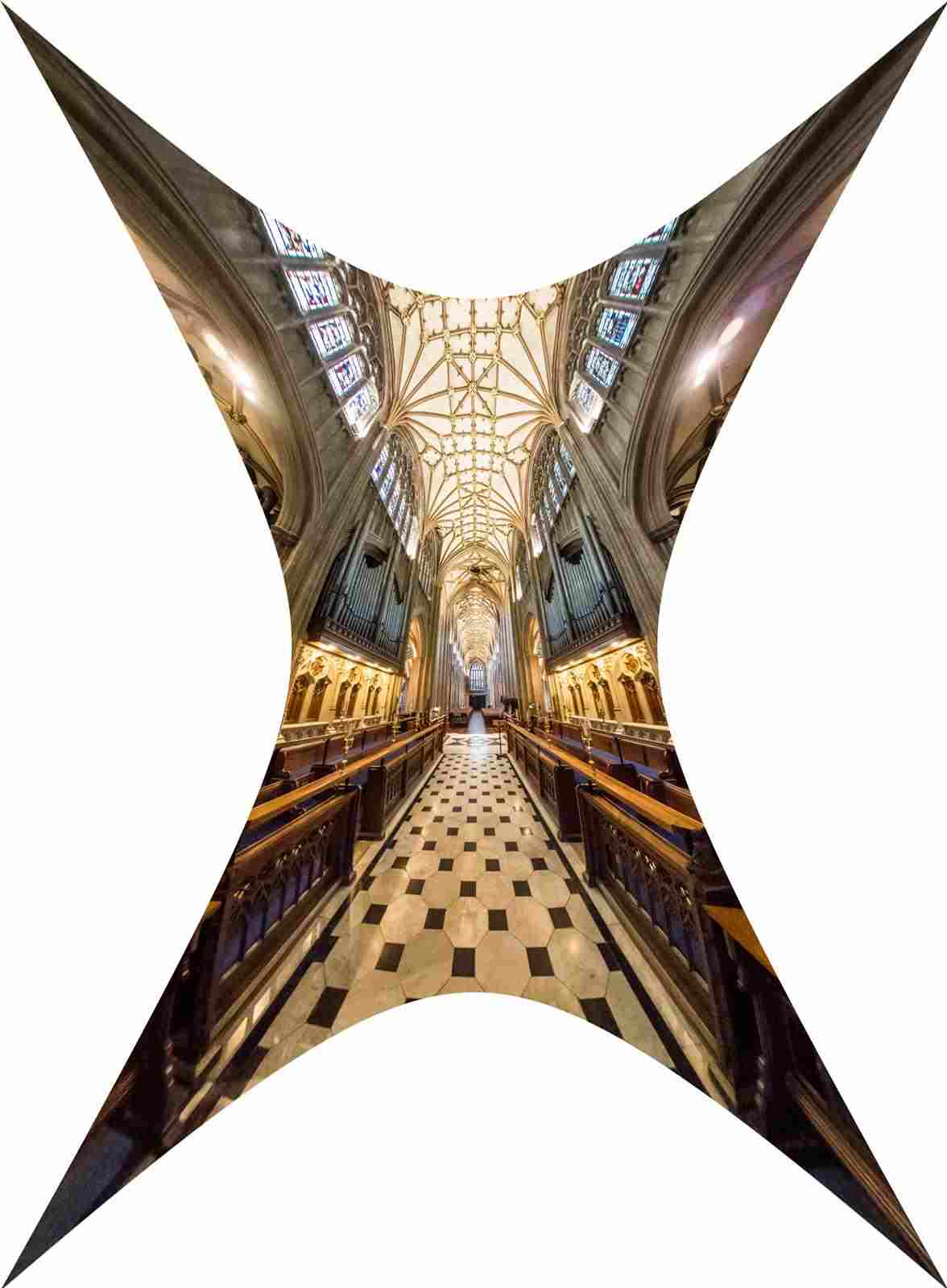}
\includegraphics[width=0.16\columnwidth]{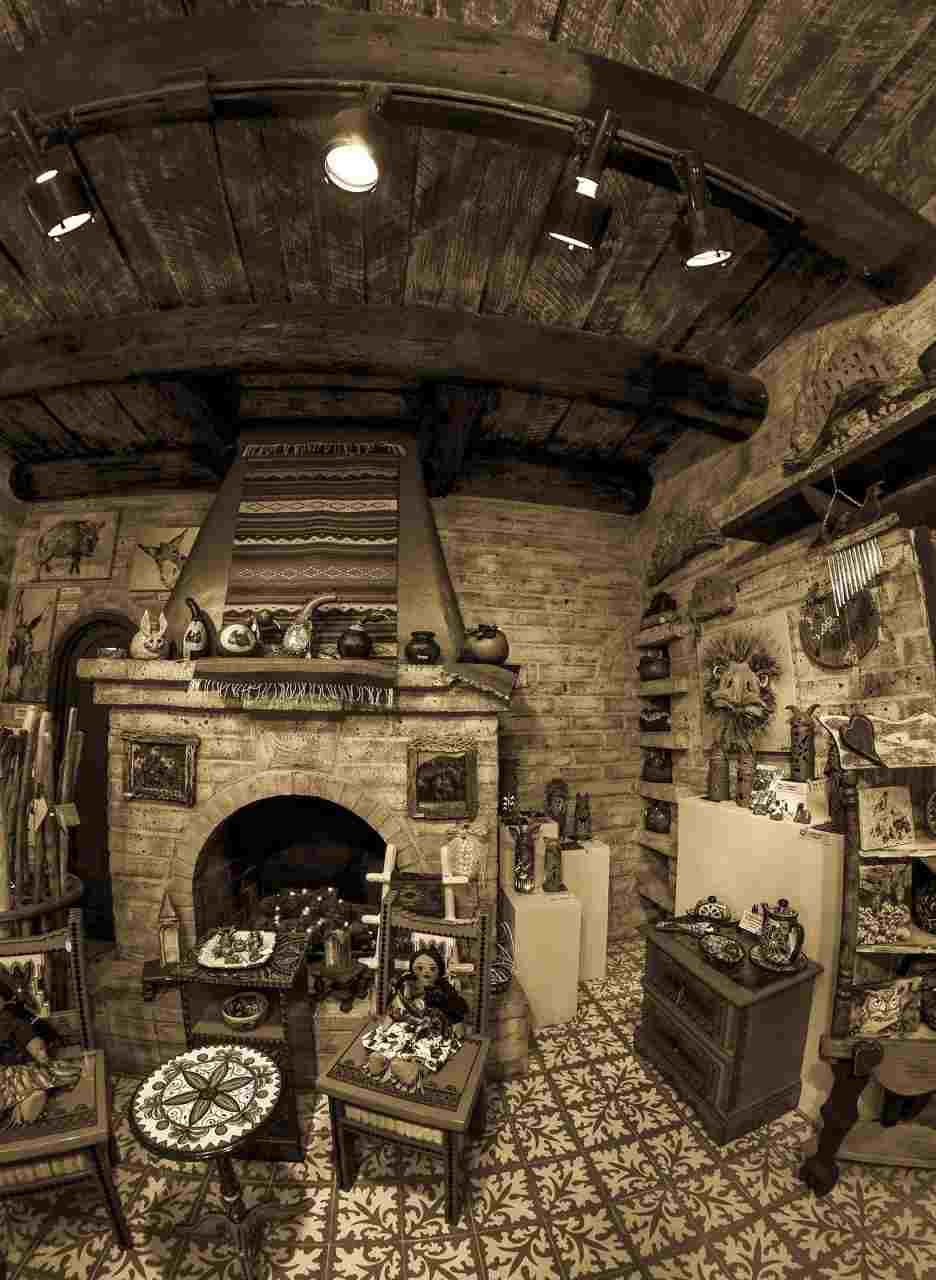}
\includegraphics[width=0.16\columnwidth]{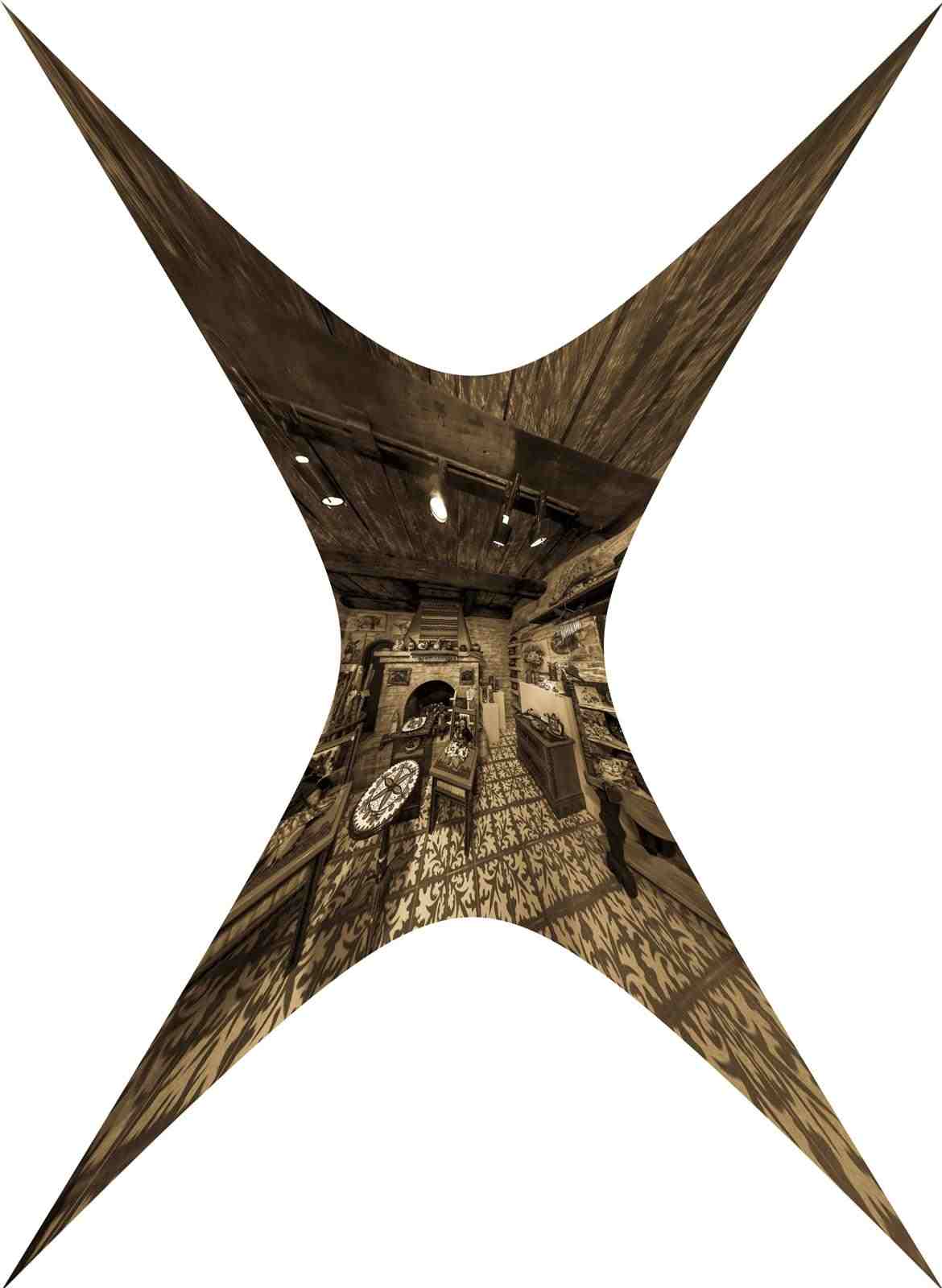}
\includegraphics[width=0.30\columnwidth]{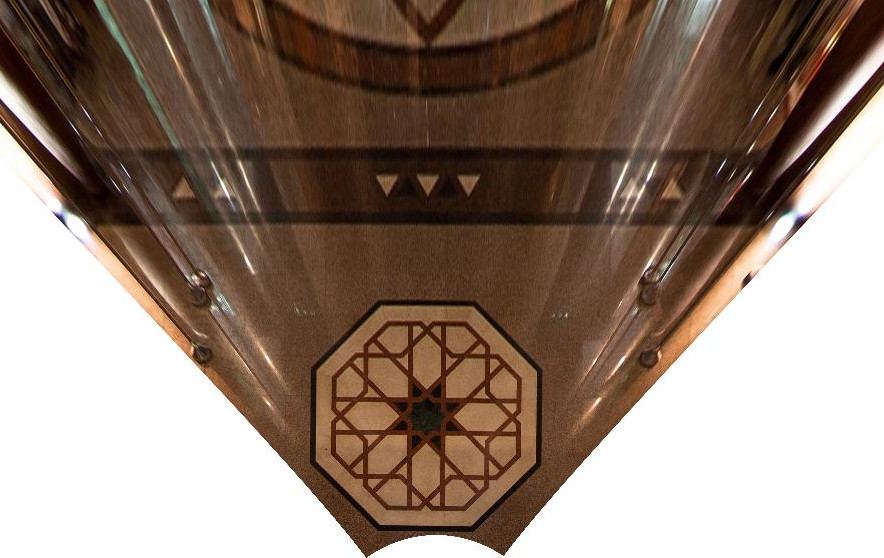}
\includegraphics[width=0.335\columnwidth]{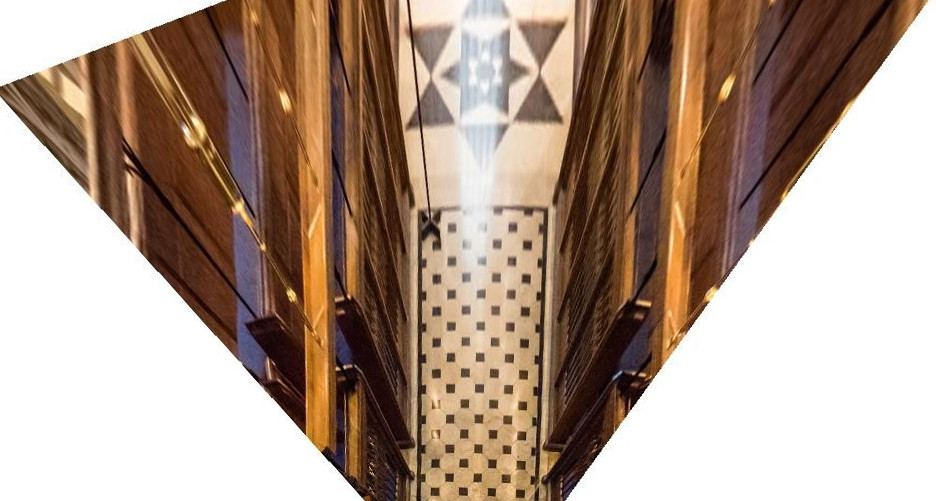}
\includegraphics[width=0.35\columnwidth]{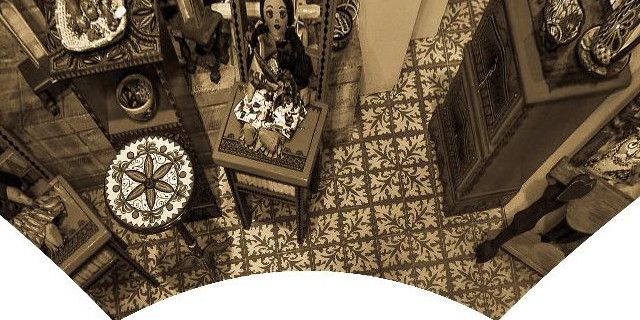}
\caption{\emph{Wide-Angle Results.} Input (top left) is an image of a
  scene plane. Outputs include the undistorted image (top right) and
  rectified scene planes (bottom row). The method is automatic.}
\label{fig:results}
\end{figure}

\label{sec:formulation}
An affine-rectifying homography \mH transforms the image of the scene
plane's vanishing line $\ve[l] = \rowvec{3}{l_1}{l_2}{l_3}^{\T}$ to
the line at infinity $\vlinf=\rowvec{3}{0}{0}{1}^{\T}$
\cite{Hartley-BOOK04}. Thus any homography $\mH$ satisfying the
constraint
\begin{equation}
  \label{eq:vline_constraint}
  \ve[l] = \ma{H}^{\T} \vlinf =
  \begin{bmatrix}  \ve[h]_1 & \ve[h]_2 & \ve[h]_3 \end{bmatrix}\colvec{3}{0}{0}{1},
\end{equation}
where \vl is an imaged scene plane's vanishing line, is an
affine-rectifying homography.  Constraint \eqref{eq:vline_constraint}
implies that $\ve[h]_3=\ve[l]$, and that the line at infinity is
invariant to rows $\ve[h]^{\T}_1$ and $\ve[h]^{\T}_2$ of \mH. Thus the
affine-rectification of image point \vx to the affine-rectified point
\vxp is defined as
\begin{equation}
\label{eq:recthg} \alpha \vxp = \mH \vx \quad \text { such that
} \mH = \begin{bmatrix} 1 & 0 & 0 \\ 0 & 1 & 0 \\ & \ve[l]^{\T}
  & \end{bmatrix}  \text{ and } \alpha \neq 0.
\end{equation}

\subsection{Radial Lens Distortion} \label{sec:radial_lens_distortion}
Rectification as given in \eqref{eq:recthg} is valid only if $\vx$ is
imaged by a pinhole camera. Cameras always have some lens distortion,
and the distortion can be significant for wide-angle lenses. For a
lens distorted point, denoted $\vxd$, an undistortion function $f$ is
needed to transform $\vxd$ to the pinhole point $\vx$. A common
parameterization for radial lens distortion is the one-parameter
division model of Fitzgibbon \etal \cite{Fitzgibbon-CVPR01}, which has
the form
\begin{equation}
\label{eq:division_model}
\vx =
f(\vxd,\lambda)=\rowvec{3}{\xd}{\yd}{1+\lambda(\xd^2+\yd^2)}^{\T},
\end{equation}
where $\vxd=\rowvec{3}{\xd}{\yd}{1}^{\T}$ is a feature point with the
distortion center subtracted. Substituting \eqref{eq:division_model}
into \eqref{eq:recthg} gives
\begin{align*}
  \alpha\vxp & = \rowvec{3}{\alpha \xp}{\alpha \yp}{\alpha}^{\T} = \mH
  \vx = \mH f(\vxd,\lambda) =
  \\ & \rowvec{3}{\xd}{\yd}{l_1\xd+l_2\yd+l_3(1+\lambda(\xd^2+\yd^2))}^{\T}. \numberthis
  \label{eq:udrect}
\end{align*}
The unknown division model parameter $\lambda$ and vanishing line \vl
appear only in the homogeneous coordinate $\alpha$. This property
simplifies the solvers derived in \Sec\ref{sec:solvers}. We also
generated a solver using the standard second-order Brown-Conrady
model~\cite{Hartley-BOOK04,Brown-PE66,Conrady-RAS19}; however, these
constraints generated a very larger solver with 85 solutions because
the radial distortion coefficients appear in the inhomogeneous
coordinates.

\section{Solvers}
\label{sec:solvers}
The proposed solvers use the invariant that rectified coplanar repeats
have equal scales. In \Secs\ref{sec:closed_form_solver}
and \ref{sec:eliminating_scales} the equal-scale invariant is used to
formulate a system of polynomial constraint equations on rectified
coplanar repeats with the vanishing line and radial distortion
parameters as unknowns. Radial lens distortion is modeled with the
one-parameter division model as defined
in \Sec\ref{sec:radial_lens_distortion}. Affine-covariant region
detections are used to model repeats since they encode the necessary
geometry for scale estimation (see \Fig\ref{fig:solver_variants}
and \Sec\ref{sec:acregions}). The solvers require 3 points from each
detected region to measure the region's scale in the image space. The
geometry of an affine-covariant region is uniquely given by an affine
frame (see \Sec\ref{sec:closed_form_solver}). Three minimal cases
exist for the joint estimation of the vanishing line and
division-model parameter (see \Fig\ref{fig:solver_variants}
and \Sec\ref{sec:eliminating_scales}). These cases differ by the
number of affine-covariant regions needed for each detected
repetition.  The method for generating the minimal solvers for the
three variants is described
in \Sec\ref{sec:creating_solvers}. Finally,
in \Sec\ref{sec:known_distortion_solver}, we show that if the
distortion parameter is given, then the constraint equations simplify,
which results in a small solver for estimating rectification under the
pinhole camera assumption.

\subsection{Equal Scales Constraint from Rectified Affine-Covariant Regions}
\label{sec:closed_form_solver}
The geometry of an oriented affine-covariant region \rgn is given by
an affine frame with its origin at the midpoint of the
affine-covariant region
detection \cite{Mikolajczyk-IJCV04,Vedaldi-SOFTWARE08}. The affine
frame is typically given as the orientation-preserving homogeneous
transformation \mA that maps from the right-handed orthonormal frame
(normalized descriptor space) to the image space as
\begin{equation*}
\label{eq:laf}
\begin{bmatrix}
\ve[y] & \ve[o] & \ve[x]
\end{bmatrix} =
\mA
\begin{bmatrix}
  0 & 0 & 1 \\ 1 & 0 & 0 \\ 1 & 1 & 1
\end{bmatrix},
\end{equation*}
where \ve[o] is the origin of the linear basis defined by \ve[x]
and \ve[y] in the image coordinate
system \cite{Mikolajczyk-IJCV04,Vedaldi-SOFTWARE08}.  Thus the matrix
$\begin{bmatrix}\ve[y] & \ve[o] & \ve[x] \end{bmatrix}$ is a
parameterization of affine-covariant region \rgn, which we call
its \emph{point-parameterization}.

Let $\begin{bmatrix} \vxd[i,1] & \vxd[i,2] & \vxd[i,3] \end{bmatrix}$
be the point parameterization of an affine-covariant region \rgnd[i]
detected in a radially-distorted image.  Then by \eqref{eq:udrect} the
affine-rectified point parameterization of \rgnd is
$\begin{bmatrix} \mH f(\vxd[i,1],\lambda) & \mH f(\vxd[i,2],\lambda)
& \mH f(\vxd[i,3],\lambda) \end{bmatrix}=\begin{bmatrix} \alpha_{i,1}\vxp[i,1]
& \alpha_{i,2}\vxp[i,2] & \alpha_{i,3}\vxp[i,3] \end{bmatrix}$, where
$\alpha_{i,j}=\ve[l]^{\T}f(\vxd[i,j],\lambda)$. Thus the
affine-rectified scale $s_i$ of \rgnd is
 \begin{align*}
 s_i &= \frac{\det \left( \begin{bmatrix} \alpha_{i,1}\vxp[i,1] & \alpha_{i,2}\vxp[i,2] & \alpha_{i,3}\vxp[i,3] \end{bmatrix}\right)}{\alpha_{i,1}\alpha_{i,2}\alpha_{i,3}} 
 = \frac{1}{\alpha_{i,1}\alpha_{i,2}\alpha_{i,3}} \cdot
 \begin{vmatrix} \xd[i,1]
 & \xd[i,2] & \xd[i,3] \\ \yd[i,1]
 & \yd[i,2] & \yd[i,3] \\ \alpha_{i,1} & \alpha_{i,2}
 & \alpha_{i,3}
 \end{vmatrix} = \\
 & \frac{\alpha_{i,1} \cdot \begin{vmatrix} \xd[i,2]
 & \xd[i,3] \\ \yd[i,2] & \yd[i,3] \\
 \end{vmatrix}
 -\alpha_{i,2} \cdot \begin{vmatrix} \xd[i,1]
 & \xd[i,3] \\ \yd[i,1] & \yd[i,3] \\
 \end{vmatrix}
 +\alpha_{i,3} \cdot \begin{vmatrix} \xd[i,1]
 & \xd[i,2] \\ \yd[i,1] & \yd[i,2] \\
 \end{vmatrix}}{\alpha_{i,1} \alpha_{i,2} \alpha_{i,3}}. \numberthis
 \label{eq:rectified_scale}
 \end{align*}
The numerator in \eqref{eq:rectified_scale} depends only on the
distortion parameter $\lambda$ and $l_3$ due to cancellations in the
determinant.  The sign of $s_i$ depends on the handedness of the
detected affine-covariant region. See \Sec\ref{sec:using_reflections}
for a method to use reflected affine-covariant regions with the
proposed solvers.

\subsection{Eliminating the Rectified Scales}
\label{sec:eliminating_scales}
The affine-rectified scale in $s_i$ \eqref{eq:rectified_scale} is a
function of the unknown undistortion parameter $\lambda$ and vanishing
line $\vl=\rowvec{3}{l_1}{l_2}{l_3}^{\T}$. A unique solution
to \eqref{eq:rectified_scale} can be defined by restricting the
vanishing line to the affine subspace $l_3=1$ or by fixing a rectified
scale, \eg, $s_1=1$. The inhomogenous representation for the vanishing
line is used since it results in degree 4 constraints in the unknowns
$\lambda, l_1,l_2$ and $s_i$ as opposed to fixing a rectified scale,
which results in complicated equations of degree 7.

Let \rgnd[i] and \rgnd[j] be repeated affine-covariant region
detections. Then the affine-rectified scales of \rgnd[i] and \rgnd[j]
are equal, namely $s_i=s_j$. Thus the unknown rectified scales of a
corresponded set of $n$ affine-covariant repeated regions
$s_1,s_2,\ldots,s_n$ can be eliminated in pairs, which gives $n-1$
algebraically independent constraints and ${n}\choose{2}$ linearly
independent equations. After eliminating the rectified scales, 3
unknowns remain, $\vl=\rowvec{3}{l_1}{l_2}{1}^{\T}$ and $\lambda$, so
3 constraints are needed. There are 3 minimal configurations for which
we derive 3 solver variants:
\begin{enumerate*}[(i)] \item 3 affine-covariant region correspondences,
which we denote as the $222$-configuration; \item 1 corresponded set
of 3 affine-covariant regions and 1 affine-covariant region
correspondence, denoted the $32$-configuration; \item and 1
corresponded set of 4 affine-covariant regions, denoted the
$4$-configuration.
\end{enumerate*} The notational convention introduced for the
input configurations---$(222,32,4)$---is extended to the bench of
state-of-the-art solvers evaluated in the experiments
(see \Sec\ref{sec:experiments}) to make comparisons between the inputs
of the solvers easy. See \Fig\ref{fig:solver_variants} for examples of
all input configurations and results from their corresponding solver
variant, and see Table~\ref{table:solver_properties} for a summary of
all tested solvers.

The system of equations is of degree 4 regardless of the input
configuration and has the form
\begin{equation}
 \label{eq:scale_constraint_eq} \alpha_{j,1}\alpha_{j,2}\alpha_{j,3}\sum_{k=1}^3(-1)^{k}M_{3,k}\alpha_{i,k}=\alpha_{i,1}\alpha_{i,2}\alpha_{i,3}\sum_{k=1}^3(-1)^{k}M_{3,k}\alpha_{j,k},
\end{equation}
where $M_{i,j}$ is the $(i,j)$-minor of matrix
$\begin{bmatrix} \alpha_{i,1} \vxp[i,1] & \alpha_{i,2}\vxp[i,2]
& \alpha_{i,3}\vxp[i,3] \end{bmatrix}$.  Note that the minors
$M_{3,\cdot}$ of \eqref{eq:scale_constraint_eq} are constant
coefficients. The $222$-configuration results in a system of 3
polynomial equations of degree 4 in three unknowns $l_1,l_2$ and
$\lambda$; the $32$-configuration results in 4 equations of degree 4,
and the $4$-configuration gives 6 equations of degree 4. Only 3
equations are needed, but we found that for the $32$- and
$4$-configurations that all ${n}\choose{2}$ equations must be used to
avoid spurious solutions that arise from vanishing $\alpha_{i,j}$ when
the rectified scales are eliminated. For example, if only equations
$s_1=s_2$, $s_1=s_3$, $s_1=s_4$ are used for the $4$-configuration
\begin{equation*}
\alpha_{i,1}\alpha_{i,2}\alpha_{i,3}
\sum_{k=1}^3(-1)^kM_{3,k}\alpha_{1,k} =
\alpha_{1,1}\alpha_{1,2}\alpha_{1,3}
\sum_{k=1}^3(-1)^kM_{3,k}\alpha_{i,k} \quad i=2,3,4\vspace{-0.15cm},
\end{equation*}
then $\lambda$ can be chosen such that
$\sum_{k=1}^3(-1)^kM_{3,k}\alpha_{1,k} = 0$, and the remaining
unknowns $l_1$ or $l_2$ chosen such that
$\alpha_{1,1}\alpha_{1,2}\alpha_{1,3} = 0$, which gives a
1-dimensional family of solutions. Thus, adding two additional
equations removes all spurious solutions.
%
%
Furthermore, including all equations simplified the elimination
template construction. In principle a solver for the
$222$-configuration can be used to solve the $32$- and
$4$-configurations. However, this decouples the scales within each
group of regions, and there will exist additional solutions that do
not satisfy the original constraints.
\subsection{Creating the Solvers}
\label{sec:creating_solvers}
We used the automatic generator from
Larsson \etal \cite{Larsson-CVPR17} to make the polynomial solvers for
the three input configurations $(222,32,4)$. The solver corresponding
to each input configuration is denoted $\mH_{222}\vl \lambda$,
$\mH_{32}\vl \lambda$, and $\mH_4\vl \lambda$, respectively. The
resulting elimination templates were of sizes $101\times 155$ (54
solutions), $107\times 152$ (45 solutions), and $115\times 151$ (36
solutions). The equations have coefficients of very different
magnitude (e.g. both image coordinates $x_i,y_i \approx10^3$ and their
squares in $x_i^2 + y_i^2 \approx 10^6$). To improve numerical
conditioning, we re-scaled both the image coordinates and the squared
distances by their average magnitudes. Note that this corresponds to a
simple re-scaling of the variables in $(\lambda,l_1,l_2)$, which is
reversed once the solutions are obtained.

Experiments on synthetic data (see \Sec\ref{sec:stability}) revealed
that using the standard \grevlex bases in the generator
of \cite{Larsson-CVPR17} gave solvers with poor numerical
stability. To generate stable solvers we used the recently proposed
basis sampling technique from
Larsson \etal \cite{Larsson-CVPR18}. In \cite{Larsson-CVPR18} the
authors propose a method for randomly sampling feasible monomial
bases, which can be used to construct polynomial solvers.  We
generated (with \cite{Larsson-CVPR17}) 1,000 solvers with different
monomial bases for each of the three variants using the heuristic
from \cite{Larsson-CVPR18}. Following the method from
Kuang \etal \cite{Kuang-ECCV12}, the sampled solvers were evaluated on
a test set of 1,000 synthetic instances, and the solvers with the
smallest median equation residual were kept. The resulting solvers
have slightly larger elimination templates ($133\times
187$,~$154\times 199$, and $115\times 151$); however, they are
significantly more stable. See \Sec\ref{sec:stability} for a
comparison between the solvers using the sampled bases and the
standard \grevlex bases (default in \cite{Larsson-CVPR17}).

\subsection{The Fixed Lens Distortion Variant}
\label{sec:known_distortion_solver}
Finally, we consider the case of known division-model parameter
$\lambda$. Fixing $\lambda$ in \eqref{eq:scale_constraint_eq} yields
degree 3 constraints in only 2 unknowns $l_1$ and $l_2$. Thus only 2
correspondences of 2 repeated affine-covariant regions are needed. The
generator of \cite{Larsson-CVPR17} found a stable solver (denoted
$\mH_{22}\vl$) with an elimination template of size $12\times 21$,
which has 9 solutions. Basis sampling was unneeded in this case.
There is second minimal problem for 3 repeated affine-covariant
regions; however, unlike the case of unknown distortion, this minimal
problem is equivalent to the $\mH_{22}$ variant. It also has 9
solutions and can be solved with the $\mH_{22}$ solver by duplicating
a region in the input. The $\mH_{22}\vl$ solver contrasts to the
solvers proposed in \cite{Ohta-IJCAI81,Criminisi-BMVC00,Chum-ACCV10}
in that it is generated from constraints directly induced by the
rectifying homography rather than its linearization.
 

\subsection{Degeneracies}
The solvers have two degeneracies. If the vanishing line passes
through the image origin $\vl=\rowvec{3}{l_1}{l_2}{0}^{\T}$, then the
radial term in the homogeneous coordinate of \eqref{eq:udrect}
vanishes. If the scene plane is fronto-parallel to the camera and
corresponding points from the affine-covariant regions fall on circles
centered at the image origin, then the radial distortion is
unobservable. The proposed solvers do not have the degeneracy of the
$\mH_{22}\ve[l]s_i$ solver of \cite{Chum-ACCV10}, which occurs if the
centroids of the sampled affine-covariant regions are colinear.

\subsection{Reflections}
\label{sec:using_reflections}
In the derivation of \eqref{eq:scale_constraint_eq}, the rectified
scales $s_i$ were eliminated with the assumption that they had equal
signs (see Sec.~\ref{sec:creating_solvers}). Reflections will have
oppositely signed rectified scales; however, reversing the orientation
of left-handed affine frames in a simple pre-processing step that
admits the use of reflections. Suppose that $\det\left(
\begin{bmatrix} \ve[\tilde{x}]_{i,1} \,
\ve[\tilde{x}]_{i,2} \, \ve[\tilde{x}]_{i,3}
\end{bmatrix}\right) < 0$, where
$(\ve[\tilde{x}]_{i,1},\ve[\tilde{x}]_{i,2},\ve[\tilde{x}]_{i,3})$ is
a distorted left-handed point parameterization of an affine-covariant
region. Then reordering the point parameterization as
$(\ve[\tilde{x}]_{i,3},\ve[\tilde{x}]_{i,2},\ve[\tilde{x}]_{i,1})$
results in a right-handed point-parameterization such that
$\det \left( \begin{bmatrix} \ve[\tilde{x}]_{i,3} \, \ve[\tilde{x}]_{i,2} \, \ve[\tilde{x}]_{i,1} \end{bmatrix}\right)
> 0$, and the scales of corresponded rectified reflections will be
equal.

\section{Robust Estimation}
The solvers are used in a \LORANSAC based robust-estimation framework
\cite{Chum-ACCV04}. Minimal samples are drawn according to the solver
variant's requirements (see \Tab\ref{table:solver_properties} and
\Fig\ref{fig:solver_variants}). Affine rectifications and
undistortions are jointly hypothesized by the solver.  A metric
upgrade is directly attempted using the minimal sample (see
\cite{Pritts-CVPR14}), and the consensus set is estimated in the
metric-rectified space by verifying the congruence of the basis
vectors of the corresponded affine frames. Models with the maximal
consensus set are locally optimized in a method similar to
\cite{Pritts-CVPR14}. The metric-rectified images are presented in the
results.

\subsection{Local Features and Descriptors}
\label{sec:acregions}
 Affine-covariant region detectors are highly repeatable on the same
 imaged scene texture with respect to significant changes of viewpoint
 and illumination \cite{Mikolajczyk-PAMI04,Mishkin-ECCV18}. Their
 proven robustness in the multi-view matching task makes them good
 candidates for representing the local geometry of repeated
 textures. In particular, we use the Maximally-Stable Extremal Region
 and Hesssian-Affine
 detectors \cite{Matas-BMVC02,Mikolajczyk-IJCV04}. The
 affine-covariant regions are given by an affine transform (see
 Sec.~\ref{sec:solvers}), equivalently 3 distinct points, which
 defines an affine frame in the image
 space \cite{Obdrzalek-BMVC02}. The image patch local to the affine
 frame is embedded into a descriptor vector by the RootSIFT
 transform \cite{Arandjelovic-CVPR12,Lowe-IJCV04}.

\subsection{Appearance Clustering and Robust Estimation}
Affine frames are tentatively labeled as repeated texture by their
appearance. The appearance of an affine frame is given by the RootSIFT
embedding of the image patch local to the affine frame.  The RootSIFT
descriptors are agglomeratively clustered, which establishes pair-wise
tentative correspondences among connected components. Each appearance
cluster has some proportion of its indices corresponding to affine
frames that represent the same repeated scene content, which are the
\emph{inliers} of that appearance cluster. The remaining affine frames
are the \emph{outliers}.

Samples for the minimal solvers are either 2 correspondences of 2
covariant regions (the $\ma{H}_{22}\cdot$ solvers), a corresponded set
of 3 covariant regions (the $\ma{H}_{32}\vl\lambda$ solver) and a
correspondence of 2 covariant regions, and a corresponded set of 4
covariant regions (the $\ma{H}_4\vl\lambda$ solver). An appearance
cluster is selected with the probability given by its relative size to
the other appearance clusters. The consensus is measured by the number
of pairs of affine frames that are mutually consistent with a rigid
transform within the appearance group, normalized by the size of each
respective group. A non-linear optimizer
following~\cite{Pritts-CVPR14} is used as the local optimization step.

\begin{table}[t!] \centering
\ra{1}
\resizebox{\textwidth}{!}{
\begin{tabular}{@{} rC{10ex}C{10ex}C{8ex}C{8ex}g{10ex}g{10ex}g{10ex}g{10ex} @{} } \toprule
& {$\mH_2\vl\vu s\lambda$} & {$\mH_{22}\vl\vu\vv s\lambda$} & {$\mH_{22}\vl s_i$} & {$\mH_{22}\lambda$} & {$\mH_{22}\vl$} & {$\mH_{222}\vl \lambda$} & {$\mH_{32}\vl \lambda$} & {$\mH_4\vl \lambda$} \\
\midrule
Reference & \cite{Pritts-CVPR18} & \cite{Pritts-CVPR18}
& \cite{Chum-ACCV10} & \cite{Fitzgibbon-CVPR01} & & & &\\ 
Rectifies & \checkmark & \checkmark & \checkmark & & \checkmark & \checkmark & \checkmark & \checkmark \\
Undistorts & \checkmark & \checkmark & & \checkmark &  & \checkmark & \checkmark & \checkmark \\
Motion & trans. & trans. & rigid & rigid\footnotemark & rigid & rigid & rigid & rigid \\
\# regions & 2 & 4 & 4 & 4  & 4 & 6  & 5  & 4 \\
\# sols.   & 2 & 4 & 1 & 18 & 9 & 54 & 45 & 36  \\
Size & 24x26 & 76x80 & 4x4 & 18x18 & 12x21 & 133x187  & 154x199  & 115x151 \\
\bottomrule
\end{tabular}
}
\caption{State-of-the-art vs. proposed (shaded in grey) solvers. The
  proposed solvers return more solutions, but typically only 1
  solution is feasible (see \Fig\ref{fig:num_sols}).}
\label{table:solver_properties}
\end{table}

\footnotetext{Correspondences must induce the same rigid transform in
  the scene plane.}
%

\section{Experiments}
\label{sec:experiments}
The stabilities and noise sensitivities of the proposed solvers are
evaluated on synthetic data. We compare the proposed solvers to a
bench of 4 state-of-the-art solvers (see
\Tab\ref{table:solver_properties}). We apply the denotations for the
solvers introduced in \Sec~\ref{sec:solvers} to all the solvers in the
benchmark, \eg, a solver requiring 2 correspondences of 2
affine-covariant regions will be prefixed by $\ma{H}_{22}$, while the
proposed solver requiring 1 corresponded set of 4 affine-covariant
regions is prefixed $\ma{H}_4$.

Included are two state-of-the-art single-view solvers for
radially-distorted conjugate translations, denoted $\mH_2\vl\vu
s\lambda$ and $\mH_{22}\vl\vu\vv s \lambda$ \cite{Pritts-CVPR18}; a
full-homography and radial distortion solver, denoted
$\mH_{22}\lambda$ \cite{Fitzgibbon-CVPR01}; and the change-of-scale
solver for affine rectification of \cite{Chum-ACCV10}, denoted
$\mH_{22}\ve[l] s_i$. The sensitivity benchmarks measure the
performance of rectification accuracy by the warp error (see
\Sec\ref{sec:warp_error}) and the relative error of the division
parameter estimate. Stability is measured with respect to the
estimated division-model parameter. The $\mH_{22}\lambda$ solver is
omitted from the warp error since the vanishing line is not estimated,
and the $\mH_{22}\ve[l] s_i$ and $\mH_{22}\ve[l]$ solvers are omitted
from benchmarks involving lens distortion since the solvers assume a
pinhole camera.

\begin{figure}[t!]
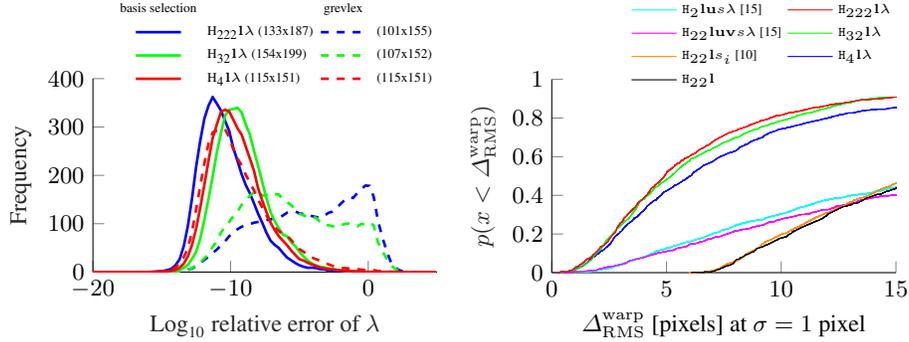

\centering
\setlength\fwidth{0.375\columnwidth} \input{fig/lambda_stability.tikz} 
\setlength\fwidth{0.375\columnwidth} \input{fig/ecdf_warp_1px_ct.tikz} 
\caption{(left) \emph{Stability study.} Shows that the basis selection method of \cite{Larsson-CVPR18} is essential for stable solver generation. (right) \emph{Proposal Study.} Reports the cumulative distributions of raw warp errors (see Sec.~\ref{sec:warp_error}) for the bench of solvers on 1000 synthetic scenes 1-$\sigma$ pixel of imaging white noise. The proposed solvers (with distortion) give significantly better proposals than the state of the art.}
\label{fig:stability_and_1pxwarp}
\end{figure}

\subsection{Synthetic Data}
\label{sec:synthetic_data}
The performance of the proposed solvers on 1000 synthetic images of 3D
scenes with known ground-truth parameters is evaluated. A camera with
a random but realistic focal length is randomly placed with respect to
a scene plane such that it is mostly in the camera's
field-of-view. The image resolution is set to 1000x1000
pixels. Conjugately translated affine frames are generated on the
scene plane such that their scale with respect to the scene plane is
realistic. The motion is restricted to conjugate translations so that
\cite{Pritts-CVPR18} can be included in the
benchmark. \Fig\ref{fig:ransac_sensitivity_study_rt} of the
supplemental includes experiments for rigidly transformed affine
frames. The modeling choice reflects the use of affine-covariant
region detectors on real images (see \Sec\ref{sec:solvers}). The image
is distorted according to the division model. For the sensitivity
experiments, isotropic white noise is added to the distorted affine
frames at increasing levels. Performance is characterized by the
relative error of the estimated distortion parameter and by the warp
error, which measures the accuracy of the affine-rectification.

\subsubsection{Warp Error}
\label{sec:warp_error}
Since the accuracy of scene-plane rectification is a primary concern,
the warp error for rectifying homographies proposed by
Pritts \etal~\cite{Pritts-BMVC16} is reported, which we extend to
incorporate the division model for radial lens
distortion \cite{Fitzgibbon-CVPR01}. A scene plane is tessellated by a
10x10 square grid of points \vXs{100} and imaged as \vxds{100} by the
lens-distorted ground-truth camera. The tessellation ensures that
error is uniformly measured over the scene plane. A round trip between
the image space and rectified space is made by
affine-rectifying \vxds{100} using the estimated division model
parameter $\hat{\lambda}$ and rectifying homography \mHhat and then
imaging the rectified plane by the ground-truth camera. Ideally, the
ground-truth camera images the rectified points onto \vxds{100}.
There is an affine ambiguity, denoted \mA, between \mHhat and the
ground-truth camera matrix \mP. The ambiguity is estimated during
computation of the warp error,
\begin{equation} 
  \label{prg:warp_residual} \Delta^{\mathrm{warp}}=\min_{\mA} \sum_{i}
  d^2(\vxd,f^d(\mP\mA\mHhat
  f(\vxd,\hat{\lambda})),\lambda),
\end{equation}
where $d(\cdot,\cdot)$ is the Euclidean distance, $f^d$ is the inverse
of the division model (the inverse of \eqref{eq:division_model}),
and \vxds{100} are the imaged grid points of the scene-plane
tessellation. The root mean square warp error for \vxds{100} is
reported and denoted as $\Delta^{\mathrm{warp}}_{\mathrm{RMS}}$.  The
vanishing line is not directly estimated by the solver
$\mH_{22}\lambda$ of \cite{Fitzgibbon-CVPR01}, so it is not reported.

\begin{figure*}[t!]
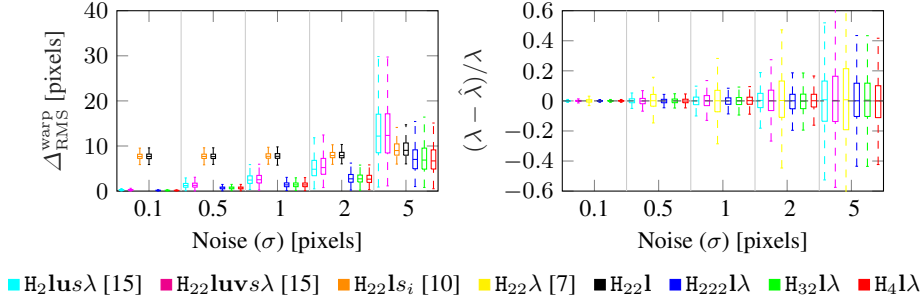

\raggedright
 \setlength\fwidth{0.35\columnwidth} \input{fig/ransac_rewarp_sensitivity_ct.tikz}
 \setlength\fwidth{0.35\columnwidth} \input{fig/ransac_rel_lambda_sensitivity_ct.tikz}
%
\centering \definecolor{mycyan}{rgb}{0,1,1}
\definecolor{myorange}{rgb}{1, 0.5490, 0}
\begin{tikzpicture}
\begin{customlegend}
[legend columns=-1,
legend style={draw=none,/tikz/every even column/.append style={column sep=0.175cm},cells={align=left}},
legend entries={$\mH_{2}\vl\vu s\lambda$ \cite{Pritts-CVPR18}, $\mH_{22}\vl\vu\vv s\lambda$ \cite{Pritts-CVPR18}, $\mH_{22}\vl s_i$ \cite{Chum-ACCV10},$\mH_{22}\lambda$ \cite{Fitzgibbon-CVPR01},$\mH_{22}\vl$,$\mH_{222}\vl \lambda$,$\mH_{32}\vl \lambda$,$\mH_4 \vl \lambda$}]
    \addlegendimage{mycyan,fill=mycyan,only marks,mark=square*}            
    \addlegendimage{magenta,fill=magenta,only marks,mark=square*}            
    \addlegendimage{myorange,fill=myorange,only marks,mark=square*}     
    \addlegendimage{yellow,fill=yellow,only marks,mark=square*}			
    \addlegendimage{black,fill=black,only marks,mark=square*}            
    \addlegendimage{blue,fill=blue,only marks,mark=square*}          
    \addlegendimage{green,fill=green,only marks,mark=square*}
    \addlegendimage{red,fill=red,only marks,mark=square*}            
\end{customlegend}
\end{tikzpicture}  
\caption{\emph{Sensitivity Benchmark.} Comparison of two error
  measures after 25 iterations of a simple \RANSAC for different
  solvers with increasing levels of white noise added to the affine
  frame correspondences. (left) Reports the warp error as
  $\Delta_{\mathrm{RMS}}^{\mathrm{warp}}$ and (right) Reports the
  relative error of the estimated division model parameter. The
  proposed solvers are significantly more robust.}
\label{fig:ransac_sensitivity_study}
\end{figure*}

\subsubsection{Numerical Stability}
\label{sec:stability}
The stability study compares the solver variants generated using the
standard \grevlex bases versus solvers generated with bases chosen by
basis sampling using \cite{Larsson-CVPR18} (see
\Sec\ref{sec:creating_solvers}). The generator of Larsson \etal
\cite{Larsson-CVPR17} \etal was used to generate both sets of
solvers. Stability is measured by the relative error of the estimated
division model parameter for noiseless affine-frame correspondences
across realistic synthetic scenes, which are generated as described in
the introduction of \Sec\ref{sec:synthetic_data}. The ground-truth
parameter of the division model $\lambda$ is drawn uniformly from the
interval $[-8,0.5]$. As a reference, the normalized division parameter
$\lambda=-4$ is typical for wide field-of-view cameras like the GoPro,
where the image is normalized by ${1}/({\text{width}+\text{height}})$.
\Fig\ref{fig:stability_and_1pxwarp} (left) reports the histogram of
$\log_{10}$ relative error of the estimates of the division model
parameter, and \Fig\ref{fig:stability_and_1pxwarp} shows that the
basis selection method of \cite{Larsson-CVPR18} significantly improves
the stability of the generated solvers. The basis-sampled solvers are
used for the remainder of the experiments.

\subsubsection{Noise Sensitivity}
\label{sec:noise_sensitivity}
The proposed and state-of-the-art solvers are tested with increasing
levels of white noise added to the point parameterizations (see
\Sec\ref{sec:closed_form_solver}) of the affine-covariant region
correspondences, which are conjugately translated (see
\cite{Pritts-CVPR18,Schaffalitzky-BMVC98}).  The amount of white noise
is given by the standard-deviation of a zero-mean isotropic Gaussian
distribution, and the solvers are tested at noise levels of $\sigma
\in \{\, 0.1,0.5,1,2,5\, \}$. The ground-truth normalized division
model parameter is set to $\lambda=-4$, which is typical for
GoPro-type imagery in normalized image coordinates.

The proposal study in the right panel of
\Fig\ref{fig:stability_and_1pxwarp} shows that for 1-pixel white
noise, the proposed solvers---$\mH_{222}
\vl\lambda$,$\mH_{32}\vl\lambda$ and $\mH_4\vl \lambda$---give
significantly more accurate estimates than the state-of-the-art
conjugate translation solvers of \cite{Pritts-CVPR18}. If 5 pixel RMS
warp error is fixed as a threshold for a good model proposal, then
50\% of the models given by the proposed solvers are good versus less
than 20\% by \cite{Pritts-CVPR18}. The proposed $\mH_{22}\vl$ solver
and $\mH_{22}\vl s_i$ \cite{Chum-ACCV10} both give biased proposals
since they don't estimate lens distortion.

For the sensitivity study in \Fig\ref{fig:ransac_sensitivity_study},
the solvers are wrapped by a basic \RANSAC estimator, which minimizes
the RMS warp error $\Delta_{\mathrm{RMS}}^{\mathrm{warp}}$ over 25
minimal samples of affine frames. The \RANSAC estimates are summarized
in boxplots for 1000 synthetic scenes. The interquartile range is
contained within the extents of a box, and the median is the
horizontal line dividing the box. As shown in
\Fig\ref{fig:ransac_sensitivity_study}, the proposed
solvers---$\mH_{222} \ve[l]\lambda,\mH_{32}\vl \lambda \text{ and }
\mH_4\vl \lambda$---give the most accurate lens distortion and
rectification estimates. The proposed solvers are superior to the
state of the art at all noise levels.  The proposed
distortion-estimating solvers give solutions with less than 5-pixel
RMS warp error $\Delta_{\mathrm{RMS}}^{\mathrm{warp}}$ 75\% of the
time and estimate the correct division model parameter more than half
the time at the 2-pixel noise level. The fixed-lens distortion solvers
$\mH_{22}\vl$ and $\mH_{22}\vl s_i$ of \cite{Chum-ACCV10} give biased
solutions since they assume the pinhole camera model.

\begin{figure}[t!]
\centering
\setlength\fwidth{0.35\columnwidth} 
%
%
\newlength\tb
\setlength\tb{0.5ex}
\begin{tikzpicture}

\begin{axis}[%
width=\fwidth,
height=0.5625\fwidth,
at={(0\fwidth,0\fwidth)},
scale only axis,
xmin=-1,
xmax=25,
xtick={0,5,10,15,20},
xlabel={Number of real solutions},
ymin=0,
ymax=4000,
ylabel style={font=\color{white!15!black}},
ylabel={Frequency},
axis background/.style={fill=white},
axis x line*=bottom,
axis y line*=left,
legend style={at={(1.025,1.05)},font=\tiny, legend cell align=left, draw=none},
enlargelimits=false
]

\addlegendentry{$\text{H}_{\text{22}}\text{\vl}\text{ (12x21)}$}
\addlegendentry{$\text{H}_{\text{222}}\text{\vl}\lambda\text{ (133x187)}$}
\addlegendentry{$\text{H}_{\text{32}}\text{\vl}\lambda\text{ (154x199)}$}
\addlegendentry{$\text{H}_\text{4}\text{\vl}\lambda\text{ (115x151)}$}

\addplot[ybar,bar width=\tb,bar shift=-0.5\tb,draw=black,fill=yellow,area legend] plot table[row sep=crcr] {%
1	3914\\
3	601\\
5	397\\
7	66\\
9	22\\
11	0\\
13	0\\
15	0\\
17	0\\
19	0\\
21	0\\
23	0\\
25	0\\
27	0\\
};
\addplot[ybar,bar width=\tb,bar shift=0.5\tb,draw=black,fill=blue,area legend] plot table[row sep=crcr] {%
1	417\\
3	1610\\
5	1283\\
7	807\\
9	454\\
11	216\\
13	113\\
15	49\\
17	26\\
19	14\\
21	4\\
23	4\\
25	3\\
27	0\\
};
\addplot[ybar,bar width=\tb,bar shift=1.5\tb,draw=black,fill=green,area legend] plot table[row sep=crcr] {%
1	199\\
3	1965\\
5	944\\
7	680\\
9	485\\
11	319\\
13	175\\
15	105\\
17	69\\
19	34\\
21	10\\
23	10\\
25	3\\
27	1\\
};
\addplot[ybar,bar width=\tb,bar shift=2.5\tb,draw=black,fill=red,area legend] plot table[row sep=crcr] {%
1	2669\\
3	388\\
5	339\\
7	388\\
9	290\\
11	315\\
13	247\\
15	147\\
17	97\\
19	74\\
21	28\\
23	13\\
25	5\\
27	0\\
};
\end{axis}
\end{tikzpicture}%
\setlength\fwidth{0.35\columnwidth} 
%
%
\begin{tikzpicture}

\begin{axis}[%
width=\fwidth,
height=0.5625\fwidth,
at={(0\fwidth,0\fwidth)},
scale only axis,
colormap={mymap}{[1pt] rgb(0pt)=(0.2422,0.1504,0.6603); rgb(1pt)=(0.25039,0.164995,0.707614); rgb(2pt)=(0.257771,0.181781,0.751138); rgb(3pt)=(0.264729,0.197757,0.795214); rgb(4pt)=(0.270648,0.214676,0.836371); rgb(5pt)=(0.275114,0.234238,0.870986); rgb(6pt)=(0.2783,0.255871,0.899071); rgb(7pt)=(0.280333,0.278233,0.9221); rgb(8pt)=(0.281338,0.300595,0.941376); rgb(9pt)=(0.281014,0.322757,0.957886); rgb(10pt)=(0.279467,0.344671,0.971676); rgb(11pt)=(0.275971,0.366681,0.982905); rgb(12pt)=(0.269914,0.3892,0.9906); rgb(13pt)=(0.260243,0.412329,0.995157); rgb(14pt)=(0.244033,0.435833,0.998833); rgb(15pt)=(0.220643,0.460257,0.997286); rgb(16pt)=(0.196333,0.484719,0.989152); rgb(17pt)=(0.183405,0.507371,0.979795); rgb(18pt)=(0.178643,0.528857,0.968157); rgb(19pt)=(0.176438,0.549905,0.952019); rgb(20pt)=(0.168743,0.570262,0.935871); rgb(21pt)=(0.154,0.5902,0.9218); rgb(22pt)=(0.146029,0.609119,0.907857); rgb(23pt)=(0.138024,0.627629,0.89729); rgb(24pt)=(0.124814,0.645929,0.888343); rgb(25pt)=(0.111252,0.6635,0.876314); rgb(26pt)=(0.0952095,0.679829,0.859781); rgb(27pt)=(0.0688714,0.694771,0.839357); rgb(28pt)=(0.0296667,0.708167,0.816333); rgb(29pt)=(0.00357143,0.720267,0.7917); rgb(30pt)=(0.00665714,0.731214,0.766014); rgb(31pt)=(0.0433286,0.741095,0.73941); rgb(32pt)=(0.0963952,0.75,0.712038); rgb(33pt)=(0.140771,0.7584,0.684157); rgb(34pt)=(0.1717,0.766962,0.655443); rgb(35pt)=(0.193767,0.775767,0.6251); rgb(36pt)=(0.216086,0.7843,0.5923); rgb(37pt)=(0.246957,0.791795,0.556743); rgb(38pt)=(0.290614,0.79729,0.518829); rgb(39pt)=(0.340643,0.8008,0.478857); rgb(40pt)=(0.3909,0.802871,0.435448); rgb(41pt)=(0.445629,0.802419,0.390919); rgb(42pt)=(0.5044,0.7993,0.348); rgb(43pt)=(0.561562,0.794233,0.304481); rgb(44pt)=(0.617395,0.787619,0.261238); rgb(45pt)=(0.671986,0.779271,0.2227); rgb(46pt)=(0.7242,0.769843,0.191029); rgb(47pt)=(0.773833,0.759805,0.16461); rgb(48pt)=(0.820314,0.749814,0.153529); rgb(49pt)=(0.863433,0.7406,0.159633); rgb(50pt)=(0.903543,0.733029,0.177414); rgb(51pt)=(0.939257,0.728786,0.209957); rgb(52pt)=(0.972757,0.729771,0.239443); rgb(53pt)=(0.995648,0.743371,0.237148); rgb(54pt)=(0.996986,0.765857,0.219943); rgb(55pt)=(0.995205,0.789252,0.202762); rgb(56pt)=(0.9892,0.813567,0.188533); rgb(57pt)=(0.978629,0.838629,0.176557); rgb(58pt)=(0.967648,0.8639,0.16429); rgb(59pt)=(0.96101,0.889019,0.153676); rgb(60pt)=(0.959671,0.913457,0.142257); rgb(61pt)=(0.962795,0.937338,0.12651); rgb(62pt)=(0.969114,0.960629,0.106362); rgb(63pt)=(0.9769,0.9839,0.0805)},
xmin=0.5,
xmax=6.5,
xtick={1, 2, 3, 4, 5, 6},
xlabel={Number of feasible solutions},
ymin=0,
ymax=5000,
ylabel={Frequency},
axis background/.style={fill=white},
axis x line*=bottom,
axis y line*=left,
legend style={legend cell align=left,align=left,draw=white!15!black}
]
\addplot[ybar,bar width=0.182,bar shift=-0.273,draw=black,fill=yellow,area legend] plot table[row sep=crcr] {%
1	3914\\
2	0\\
3	601\\
4	0\\
5	397\\
6	0\\
};
\addplot[ybar,bar width=0.182,bar shift=-0.091,draw=black,fill=blue,area legend] plot table[row sep=crcr] {%
1	4846\\
2	97\\
3	43\\
4	5\\
5	6\\
6	1\\
};
\addplot[ybar,bar width=0.182,bar shift=0.091,draw=black,fill=green,area legend] plot table[row sep=crcr] {%
1	4836\\
2	100\\
3	40\\
4	11\\
5	7\\
6	0\\
};
\addplot[ybar,bar width=0.182,bar shift=0.273,draw=black,fill=red,area legend] plot table[row sep=crcr] {%
1	4802\\
2	100\\
3	46\\
4	22\\
5	8\\
6	14\\
};
\end{axis}
\end{tikzpicture}%
\caption{(left) \emph{Real Solutions.} The histogram of real solutions for the proposed solvers. (right) \emph{Feasible Solutions.} There is typically only 1 feasible real solution.}
\label{fig:num_sols}
\end{figure}
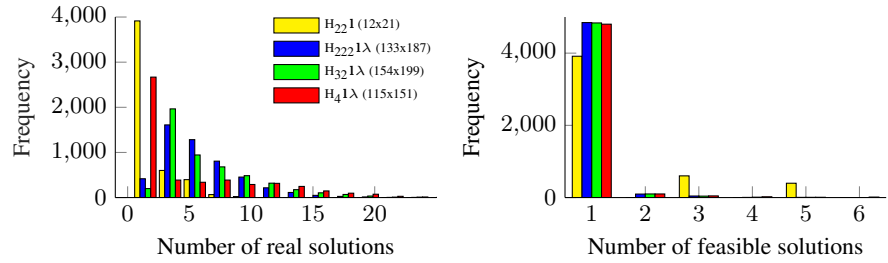

\subsubsection{Feasible Solutions and Runtime}
\Fig\ref{fig:num_sols} (left) shows the number of real solutions given
by the proposed solvers for 5000 synthetic scenes, and
\Fig\ref{fig:num_sols} (right) shows the subset of feasible solutions
as defined by the estimated normalized division-model parameter
solution falling in the interval $[-8,0.5]$. All solutions are
considered feasible for the $\mH_{22}\vl$
solver. \Fig\ref{fig:num_sols} (right) shows that in $97 \%$ of the
scenes only 1 solution is feasible, which means that nearly all
incorrect solutions can be quickly discarded.  The runtimes of the
MATLAB implementation of the solvers on a standard desktop are 2 ms
for {$\mH_{222}\vl \lambda$}, 2.2 ms for {$\mH_{32}\vl \lambda$}, 1.7
ms for {$\mH_4\vl \lambda$}, and 0.2 ms for {$\mH_{22}\vl$}.

\subsection{Real Images}
\begin{figure}[t!]
\includegraphics[width=0.28\columnwidth]{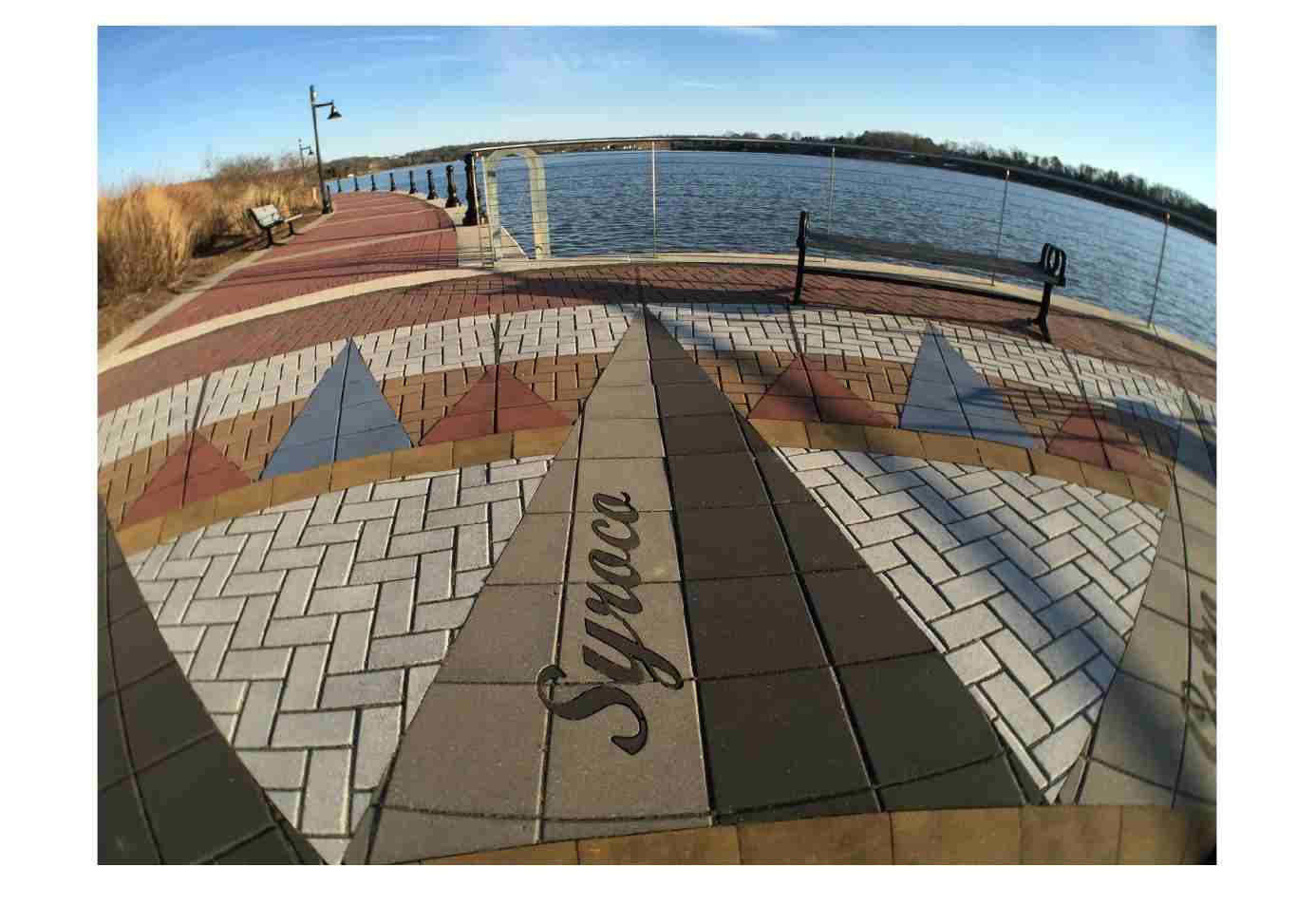}
\includegraphics[width=0.22\columnwidth]{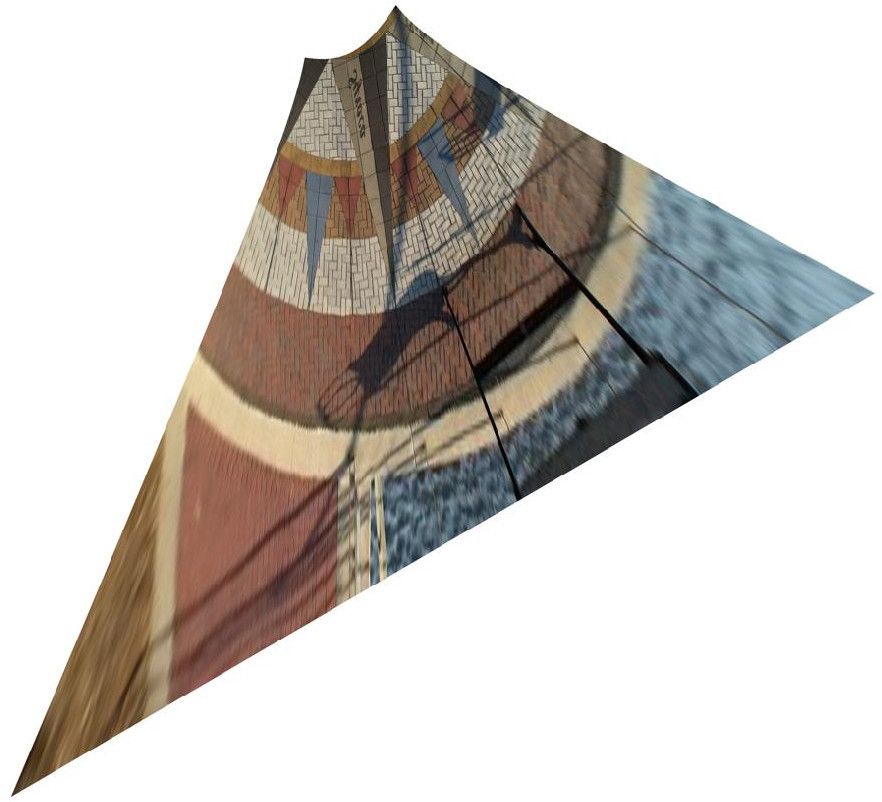}
\includegraphics[width=0.25\columnwidth]{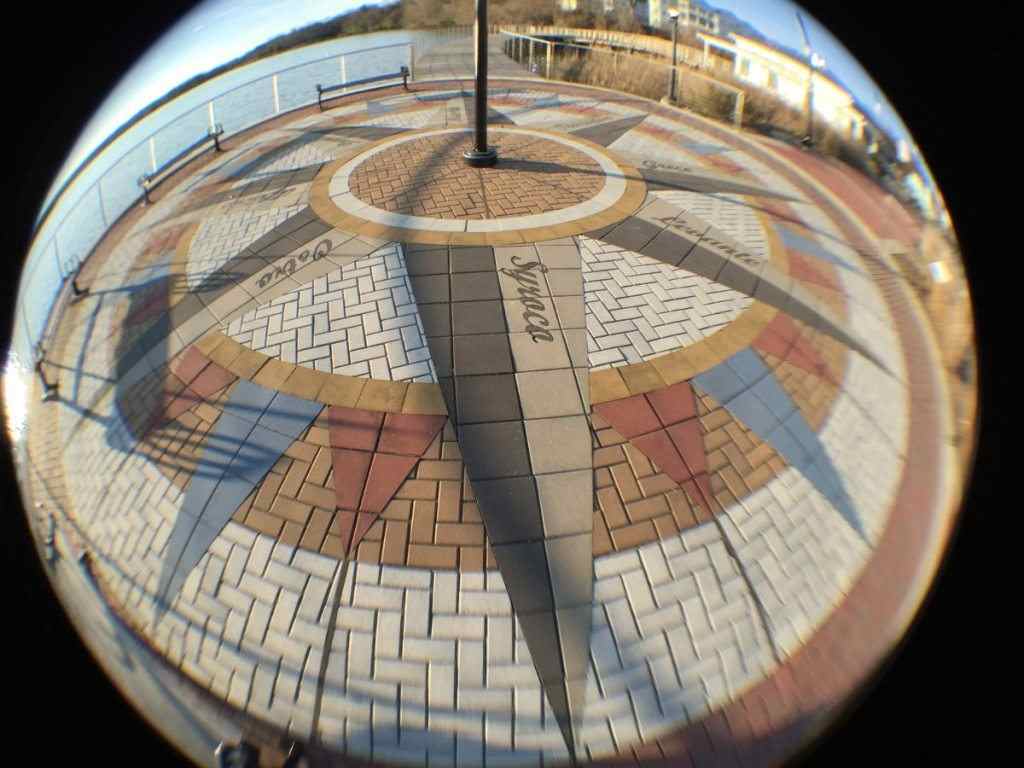}
\includegraphics[width=0.18\columnwidth]{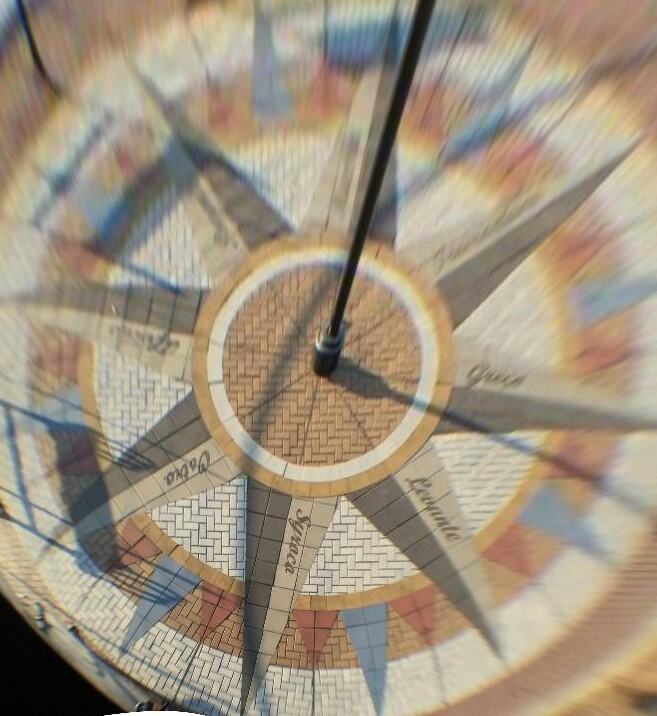}
\caption{(left pair) The waterfront is a circle, which violates the
  plumb-line assumption. (right pair) A good rectification is
  estmiated by the proposed method even with a fish-eye lens.}
\label{fig:challenging_img}
\end{figure}

\begin{figure}[H]
\begin{minipage}{0.24\textwidth}
\centering
[narrow]
\includegraphics[width=\textwidth]{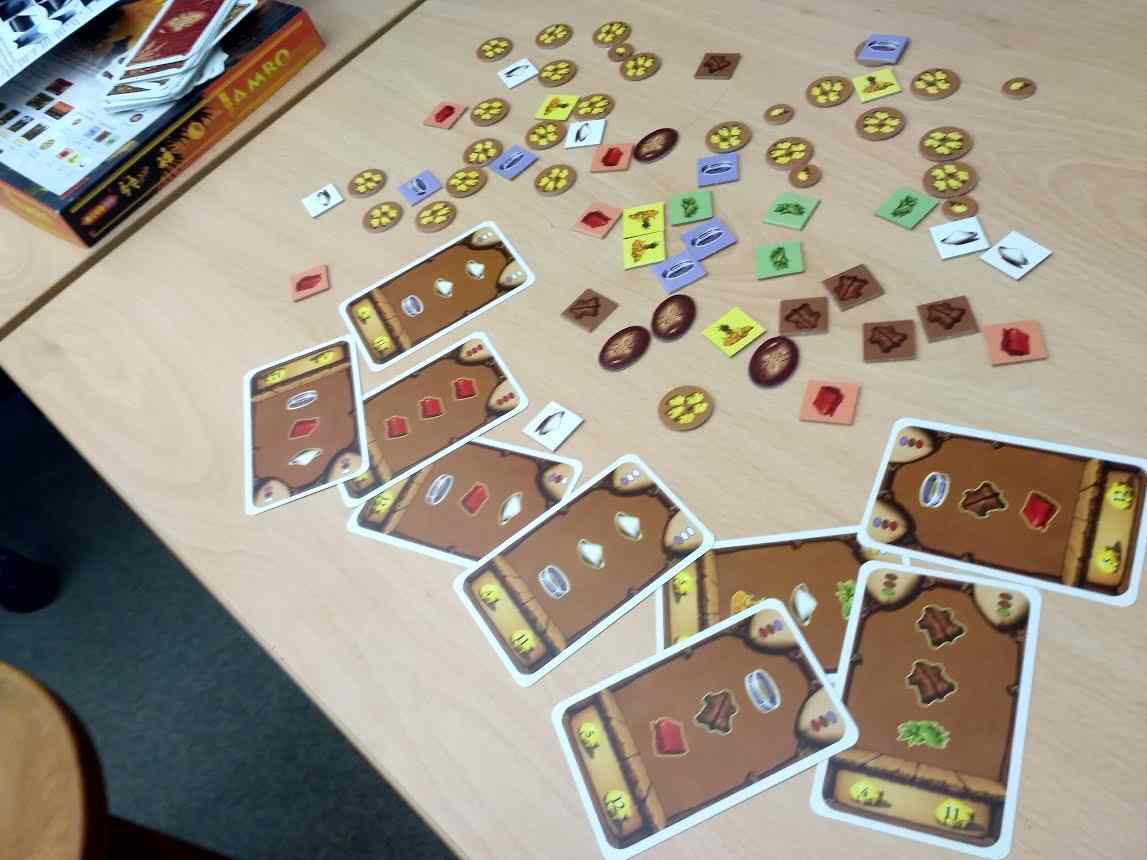}
\end{minipage}
\begin{minipage}{0.24\textwidth}
\centering
[medium]
\includegraphics[width=\textwidth]{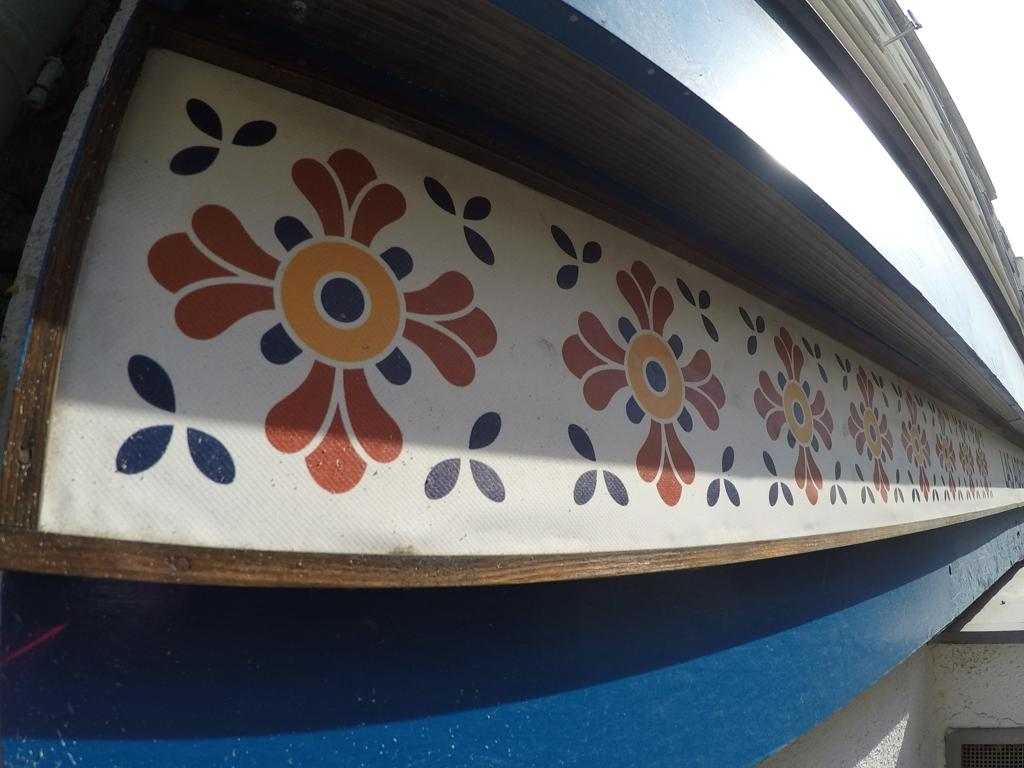}
\end{minipage}
\begin{minipage}{0.24\textwidth}
\centering
[wide]
\includegraphics[width=\textwidth]{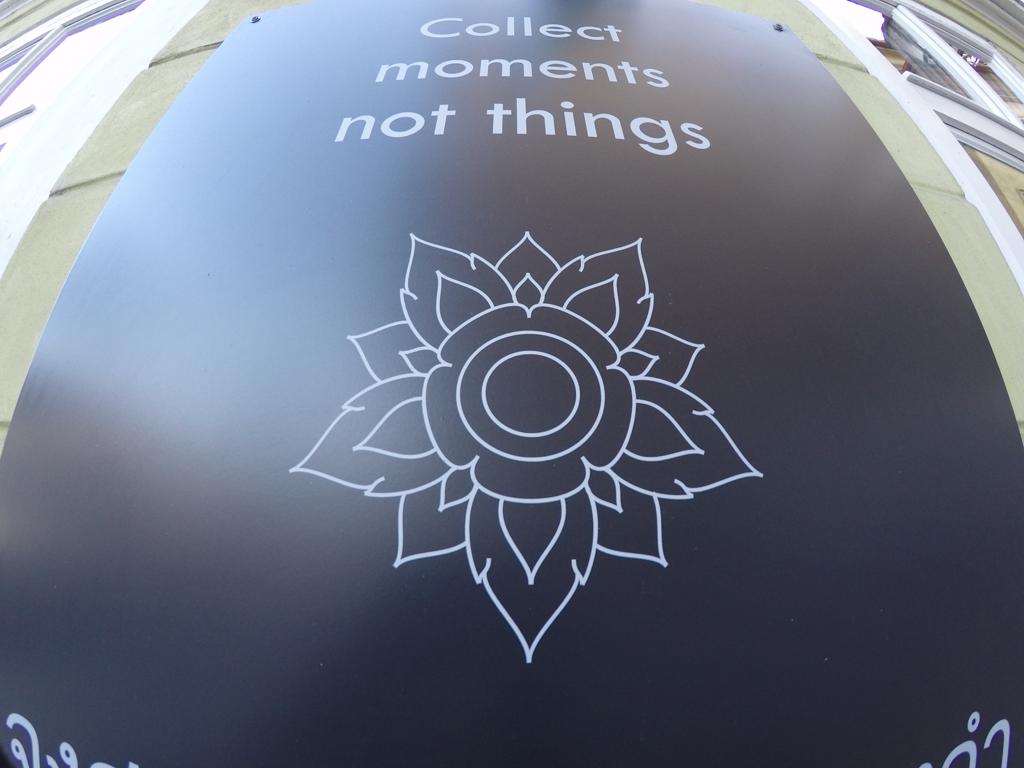}
\end{minipage}
\begin{minipage}{0.24\textwidth}
\centering
[fisheye]
\includegraphics[width=\textwidth]{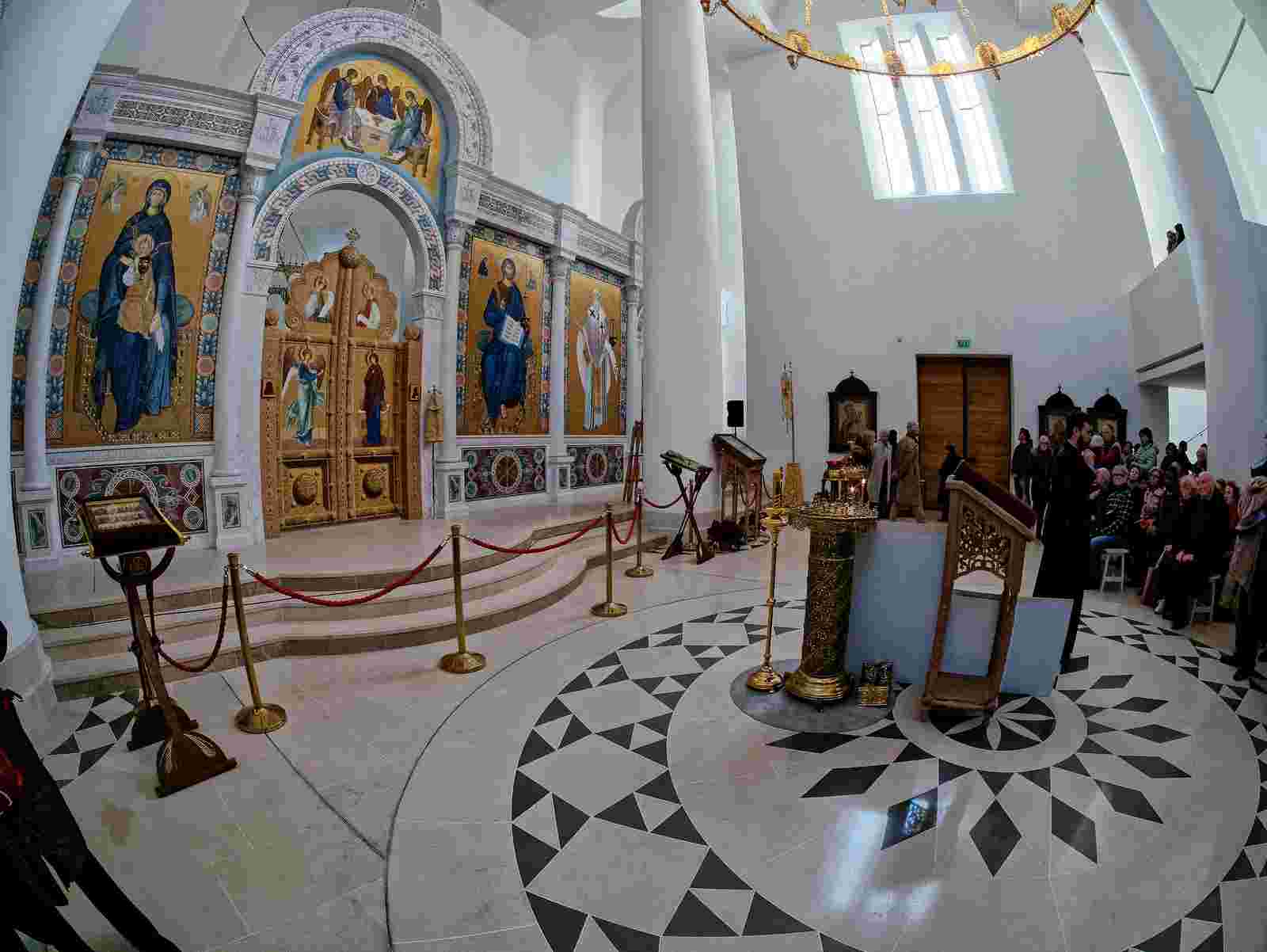}
\end{minipage}
\begin{minipage}{0.24\textwidth}
\centering
\includegraphics[width=\textwidth]{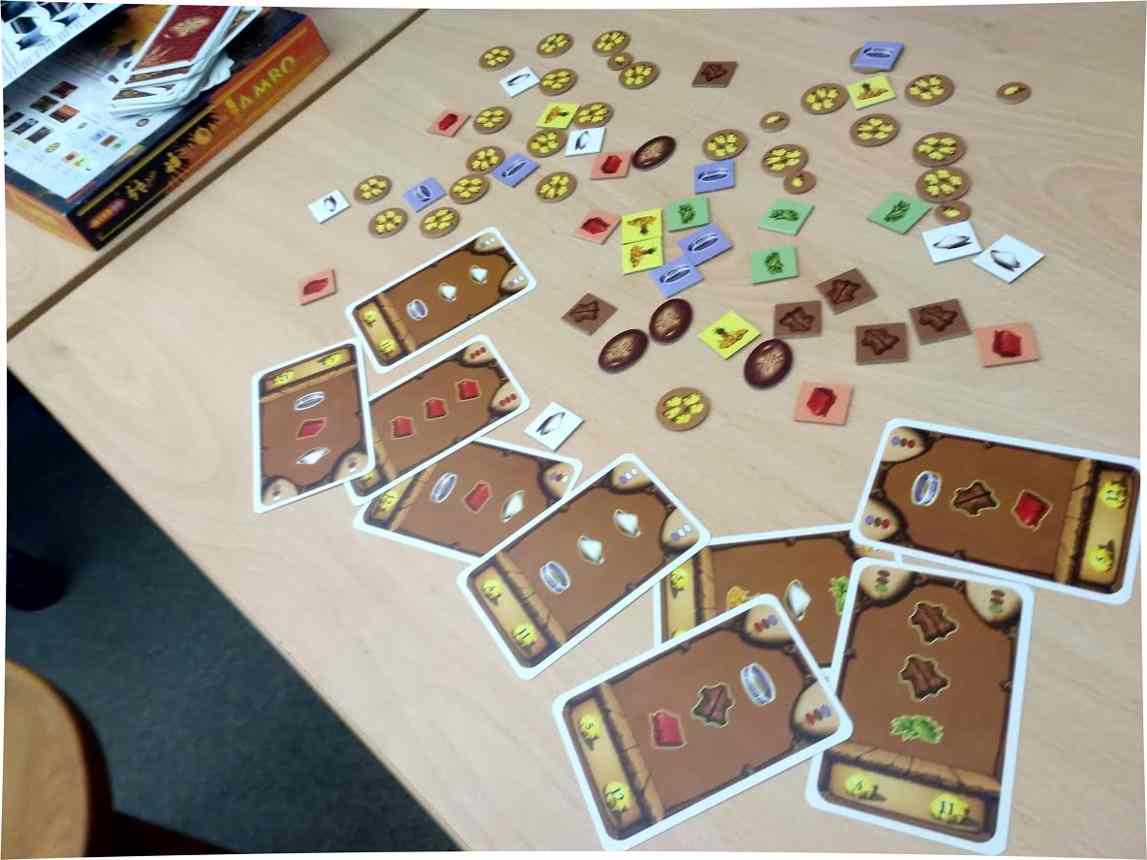}
\end{minipage}
\begin{minipage}{0.24\textwidth}
\centering
\includegraphics[width=\textwidth]{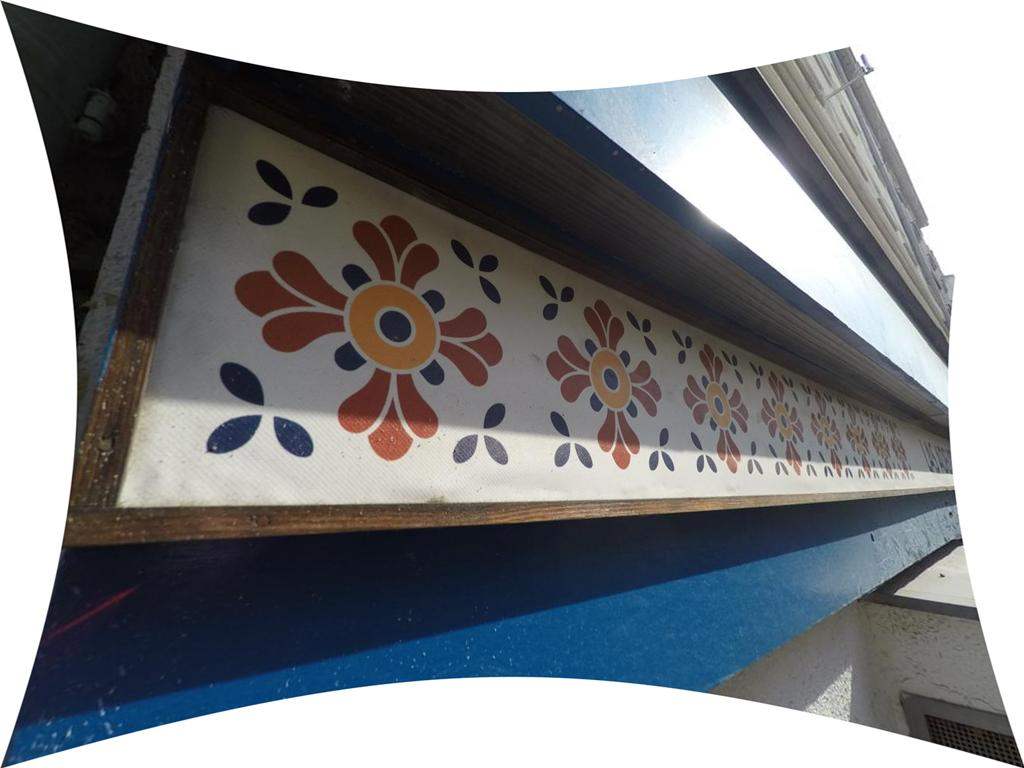}
\end{minipage}
\begin{minipage}{0.24\textwidth}
\centering
\includegraphics[width=\textwidth]{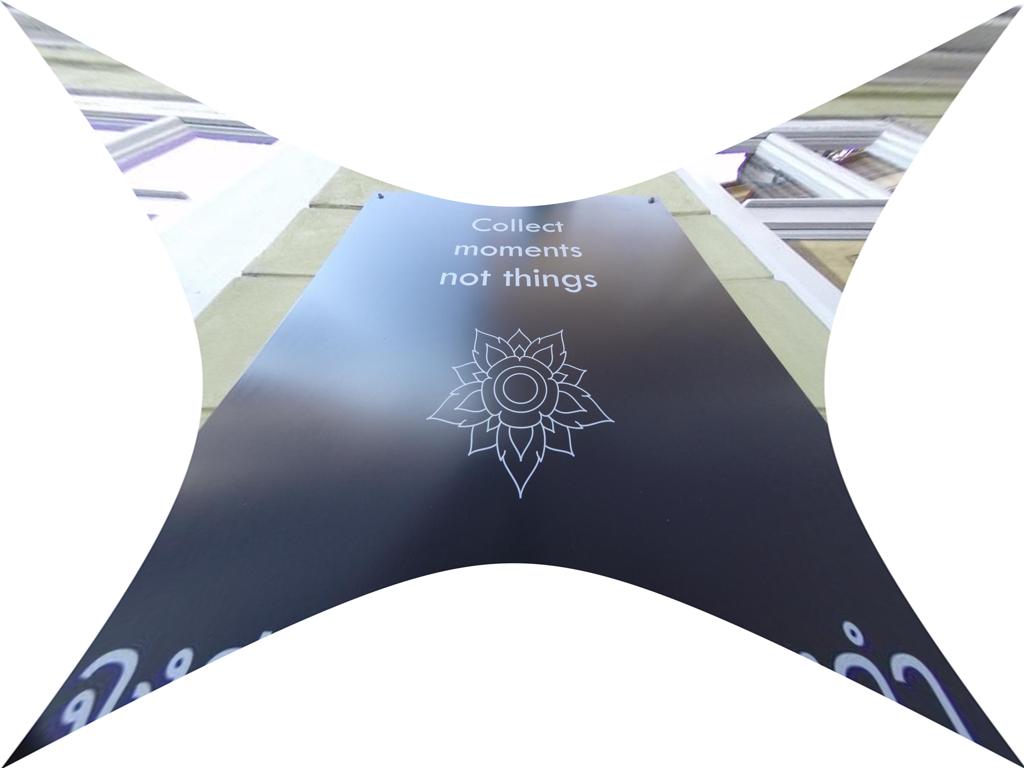}
\end{minipage}
\begin{minipage}{0.24\textwidth}
\centering
\includegraphics[width=\textwidth]{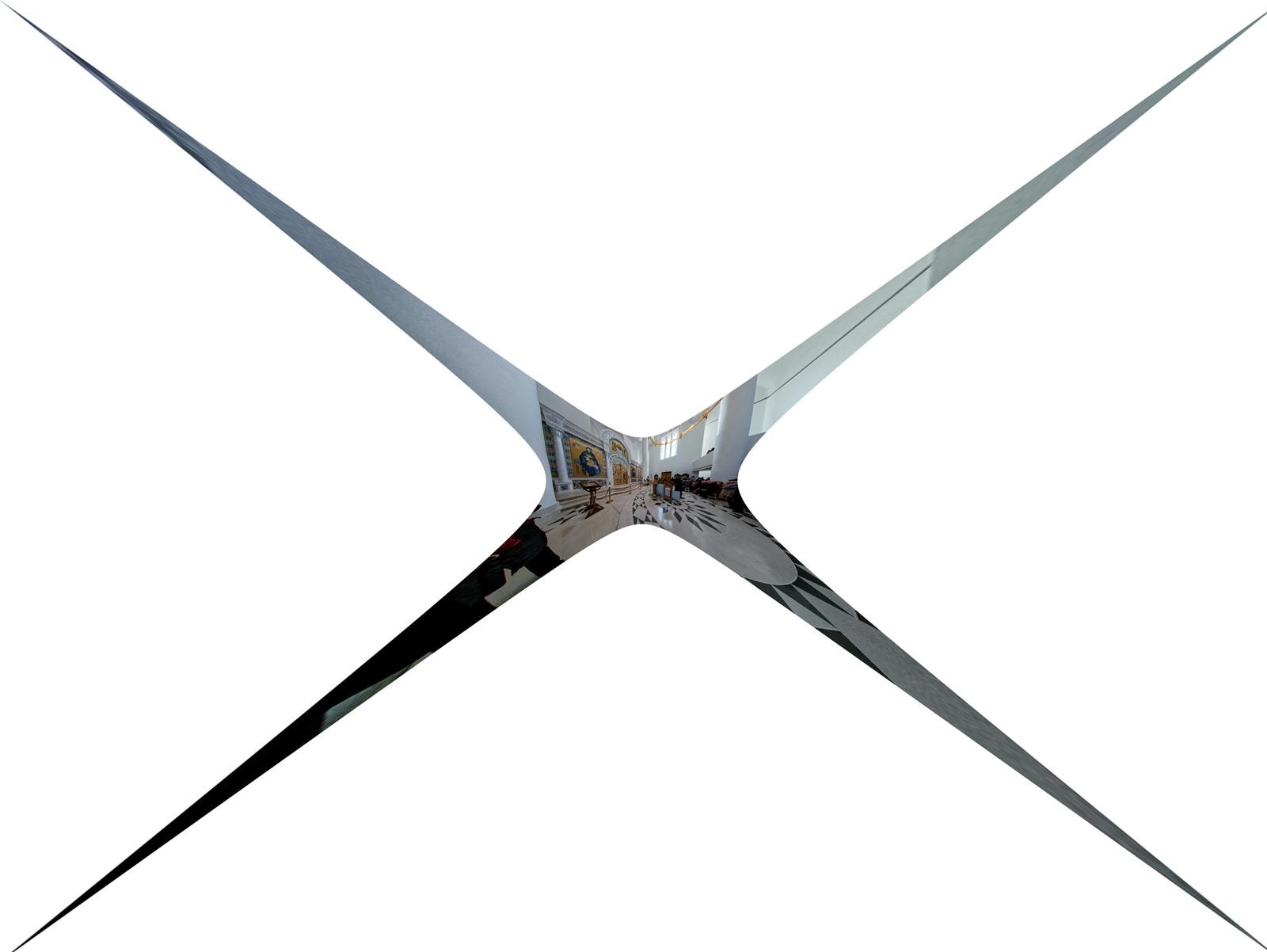}
\end{minipage}
\begin{minipage}{0.24\textwidth}
\centering
\includegraphics[width=\textwidth]{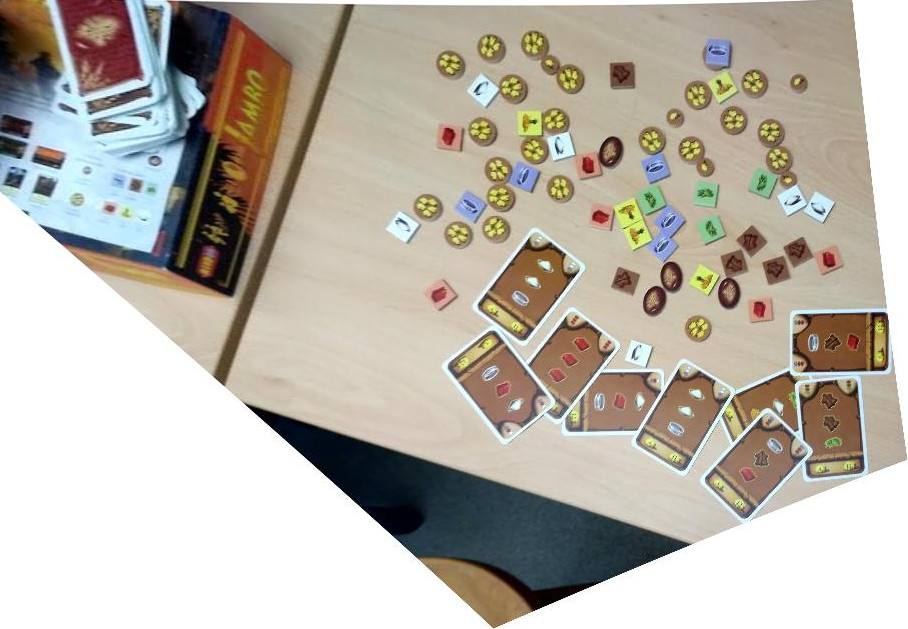}
\end{minipage}
\begin{minipage}{0.24\textwidth}
\centering
\includegraphics[width=\textwidth]{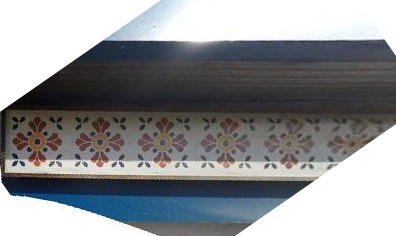}
\end{minipage}
\begin{minipage}{0.24\textwidth}
\centering
\includegraphics[width=\textwidth]{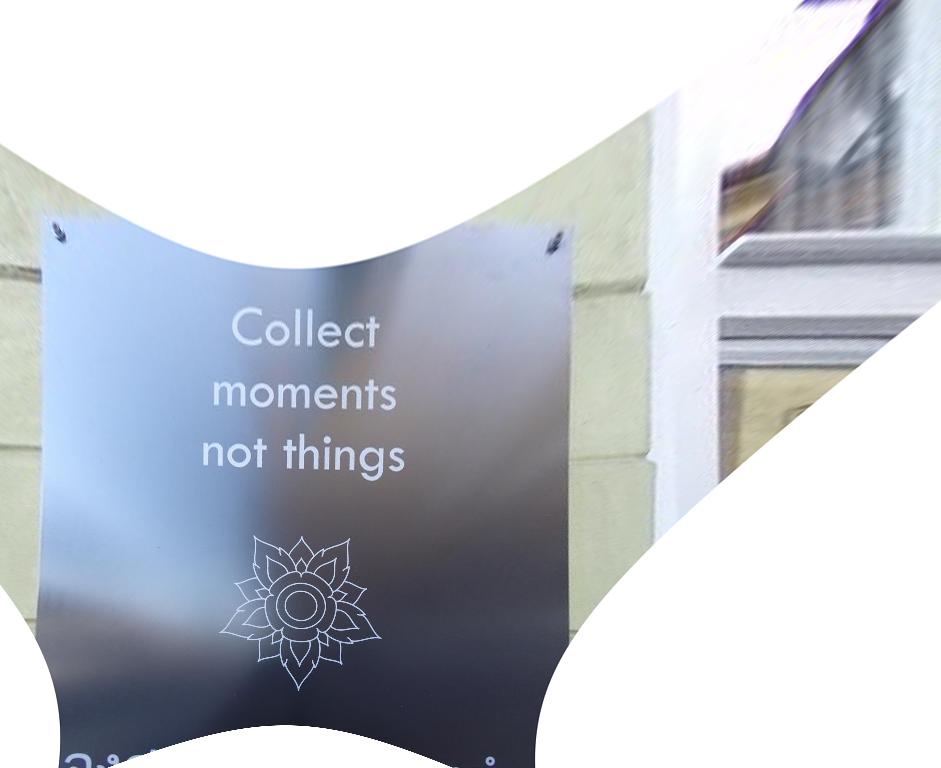}
\end{minipage}
\begin{minipage}{0.24\textwidth}
\centering
\includegraphics[width=\textwidth]{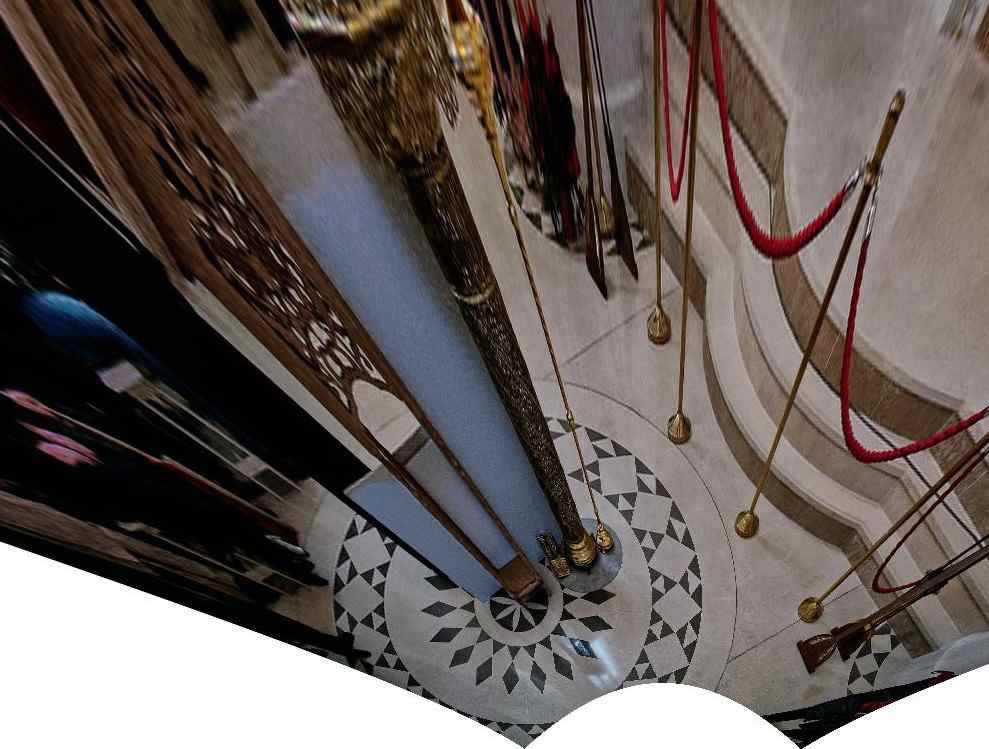}
\end{minipage}
\caption{\emph{Field-of-View Study.} The proposed solver $\mH_{222}\vl\lambda$ gives
accurate rectifications across all fields-of-view: (left-to-right)
Android phone, GoPro Hero 4 at the medium- and wide-FOV settings, and
a Panasonic DMC-GM5 with a Samyang 7.5mm fisheye lens. The outputs are
the undistorted (middle row) and rectified images (bottom row).}
\label{fig:field_of_view}
\end{figure}

The field-of-view experiment of \Fig\ref{fig:field_of_view} evaluates
the proposed $\mH_{222}\vl\lambda$ solver on real images taken with
narrow, medium, wide-angle, and fish-eye lenses. Images with diverse
scene content were chosen. \Fig\ref{fig:field_of_view} shows that the
$\mH_{222}\vl\lambda$ gives accurate rectifications for all lens
types. Additional results for wide-angle lenses are included in
\Sec\ref{sec:extended_experiments} of the
supplemental. \Fig\ref{fig:solver_comparison} compares the proposed
$\mH_{222}\vl\lambda$ and $\mH_{22}\vl$ solvers to the
state-of-the-art solvers on images with increasing levels of radial
lens distortion (top to bottom) that contain either translated or
rigidly-transformed coplanar repeated patterns. Only the proposed
$\mH_{222}\vl\lambda$ accurately rectifies on both pattern types and
at all levels of distortion. The results are after a local
optimization and demonstrate that the method of
Pritts~\etal~\cite{Pritts-CVPR14} is unable to accurately rectify
without a good initial guess at the lens distortion.  The proposed
fixed-distortion solver $\mH_{22}\vl$ gave a better rectification than
the change-of-scale solver of Chum \etal
\cite{Chum-ACCV10}. \Fig\ref{fig:challenging_img} shows the
rectifications of a deceiving picture of a landmark taken by
wide-angle and fisheye lenses. From the wide-angle image it is not
obvious which lines are really straight in the scene making
undistortion with the plumb-line constraint difficult.

\section{Conclusion}
This paper proposes solvers that extend affine-rectification to
radially-distorted images that contain essentially arbitrarily
repeating coplanar patterns. Synthetic experiments show that the
proposed solvers are more robust to noise with respect to the
state of the art while being applicable to a broader set of image
content. The paper demonstrates that robust solvers can be generated
with by the basis selection method of \cite{Larsson-CVPR18} by
maximizing for numerical stability. Experiments on difficult images
with large radial distortions confirm that the solvers give
high-accuracy rectifications if used inside a robust estimator. By
jointly estimating rectification and radial distortion, the proposed
minimal solvers eliminate the need for sampling lens distortion
parameters in \RANSAC.

\subsubsection{Acknowledgements}
James Pritts acknowledges the European Regional Development Fund under
the project Robotics for Industry 4.0
(reg. no. CZ.02.1.01/0.0/0.0/15\_003/0000470) and grants MSMT LL1303
ERC-CZ and SGS17/185/OHK3/3T/13; Zuzana Kukelova the ESI Fund, OP RDE
programme under the project International Mobility of Researchers
MSCA-IF at CTU No. CZ.02.2.69/0.0/0.0/17\_050/0008025;
and Ondrej Chum grant MSMT LL1303 ERC-CZ.

\begin{figure*}
\begin{minipage}{0.01\textwidth}
\hfill
\centering
\end{minipage}
\begin{minipage}{0.2\textwidth}
\centering
[$\mH_{22}\ve[l]\ve[u]\ve[v]\lambda$ + LO]
\includegraphics[width=\textwidth]{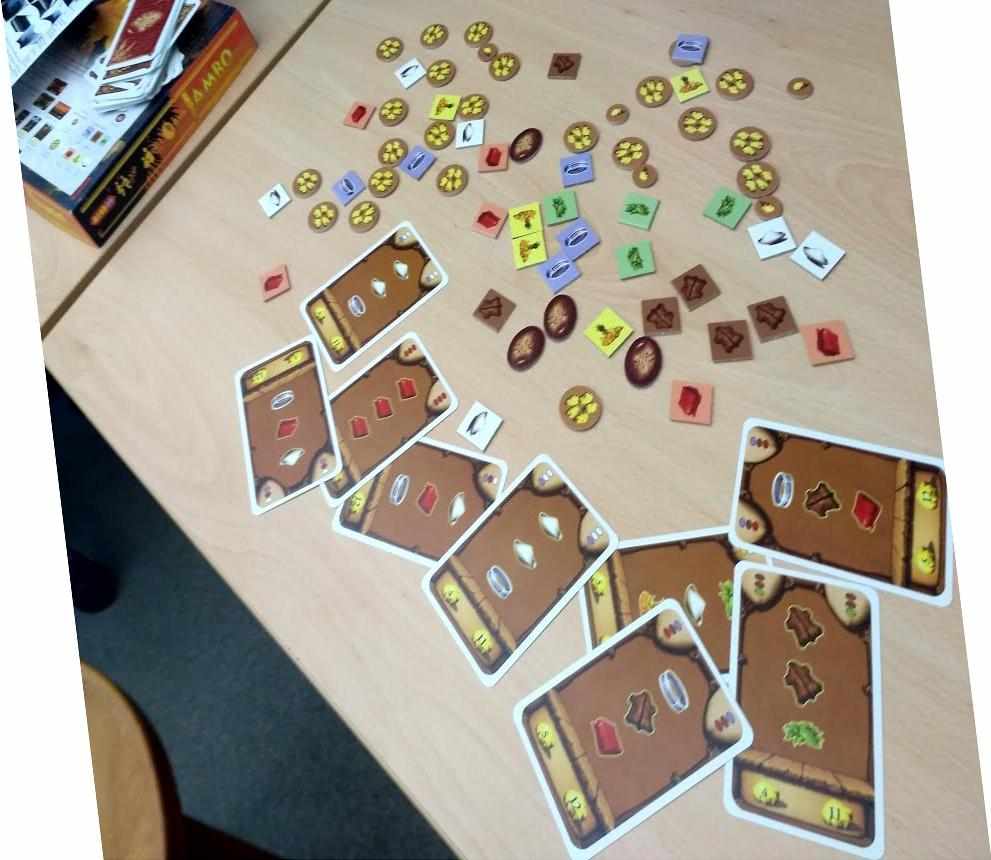}
\end{minipage}
\begin{minipage}{0.01\textwidth}
\hfill
\centering
\end{minipage}
\begin{minipage}{0.225\textwidth}
\centering
[$\mH_{22}\vl s_i$ + LO]
\includegraphics[width=\textwidth]{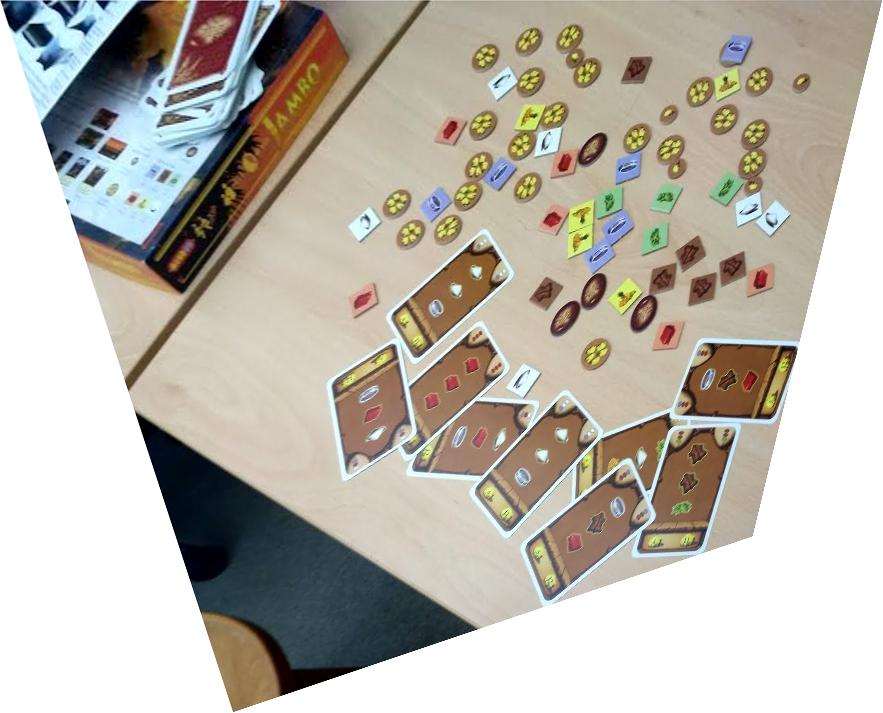}
\end{minipage}
\begin{minipage}{0.245\textwidth}
\centering
[$\mH_{22}\vl$ + LO]
\includegraphics[width=\textwidth]{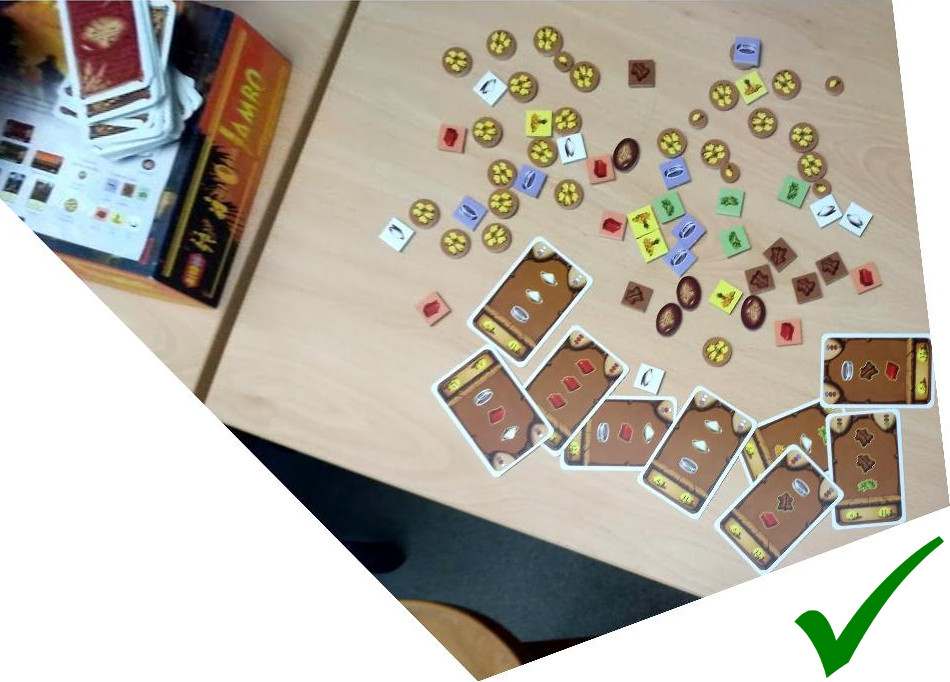}
\end{minipage}
\begin{minipage}{0.005\textwidth}
\hfill
\centering
\end{minipage}
\begin{minipage}{0.245\textwidth}
\centering
[$\mH_{222}\vl \lambda$ + LO]
\includegraphics[width=\textwidth]{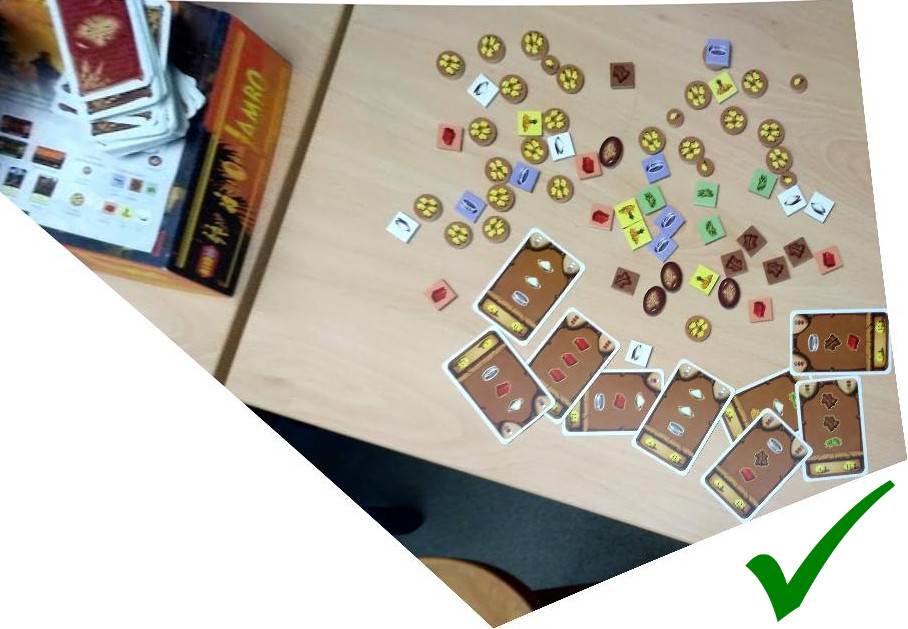}
\end{minipage}
\begin{minipage}{0.01\textwidth}
\hfill
\centering
\end{minipage}
\begin{minipage}{0.245\textwidth}
\centering
\includegraphics[width=\textwidth]{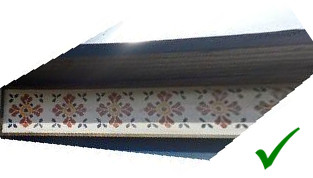}
\end{minipage}
\begin{minipage}{0.22\textwidth}
\centering
\includegraphics[width=\textwidth]{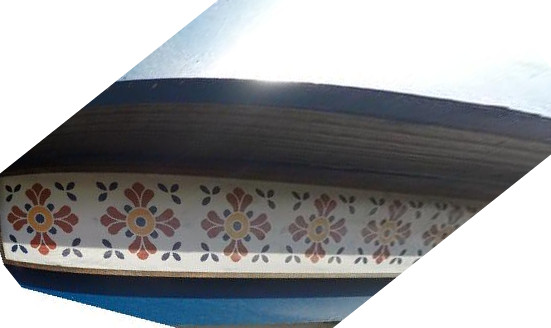}
\end{minipage}
\begin{minipage}{0.245\textwidth}
\centering
\includegraphics[width=\textwidth]{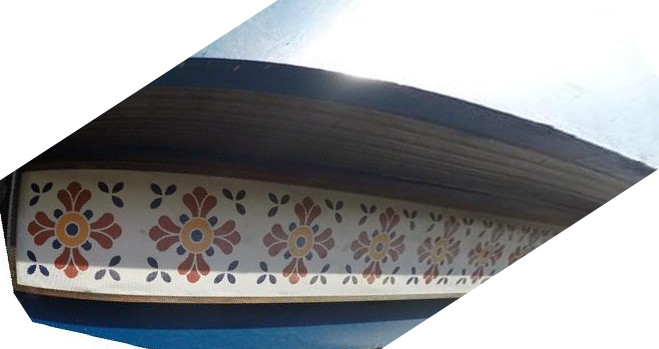}
\end{minipage}
\begin{minipage}{0.245\textwidth}
\centering
\includegraphics[width=\textwidth]{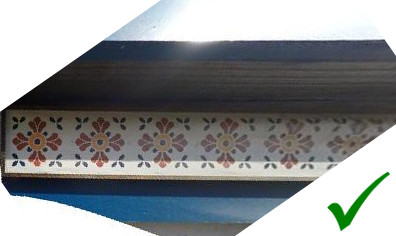}
\end{minipage}
\begin{minipage}{0.245\textwidth}
\centering
\includegraphics[width=\textwidth]{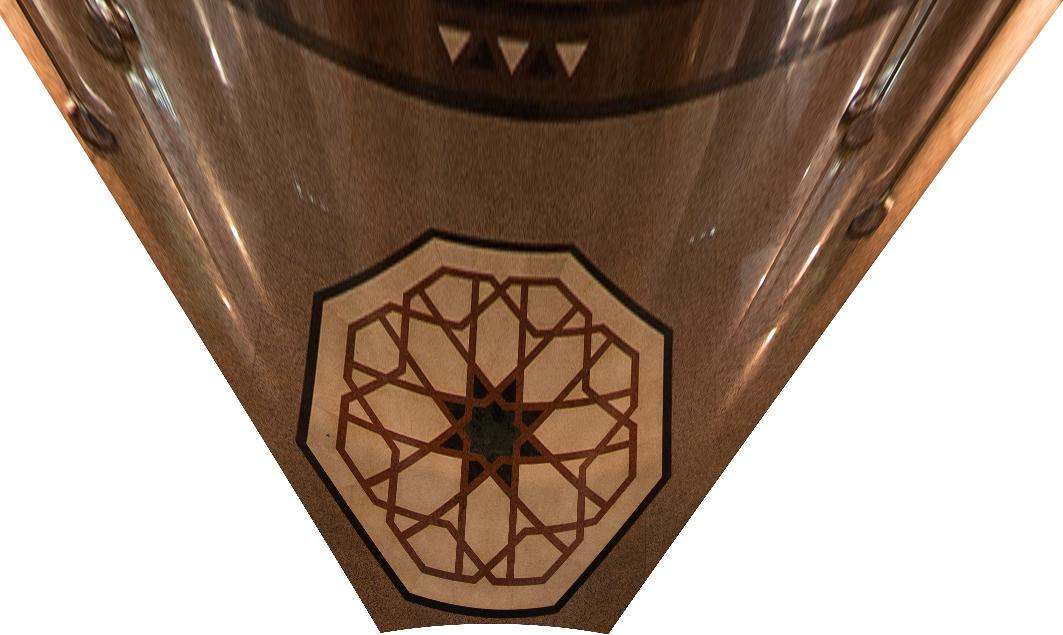}
\end{minipage}
\begin{minipage}{0.22\textwidth}
\centering
\includegraphics[width=\textwidth]{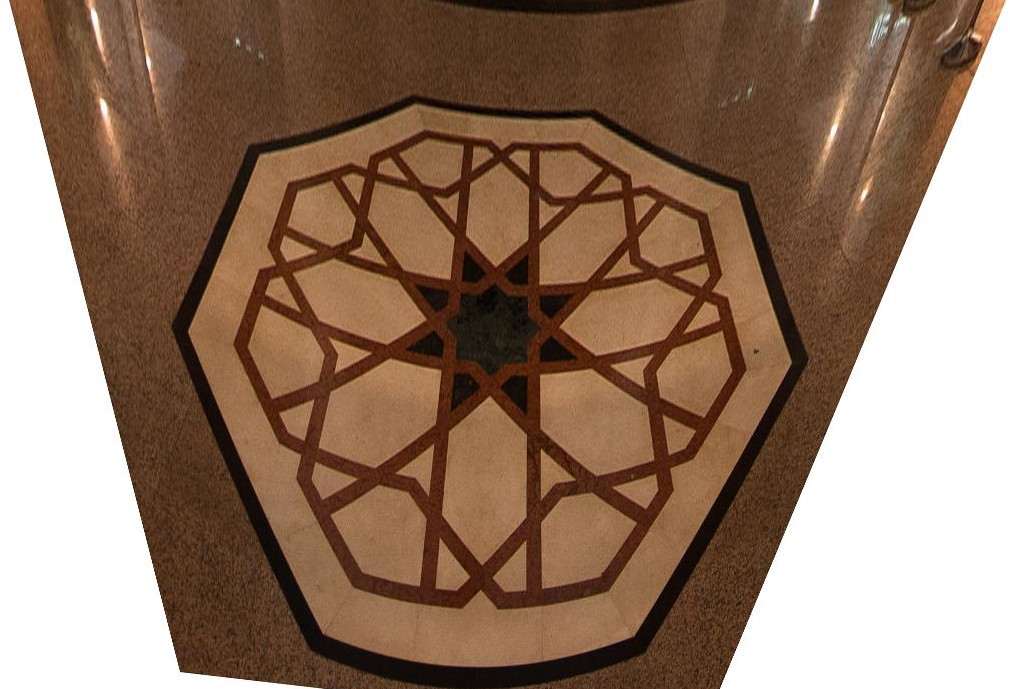}
\end{minipage}
\begin{minipage}{0.245\textwidth}
\centering
\includegraphics[width=\textwidth]{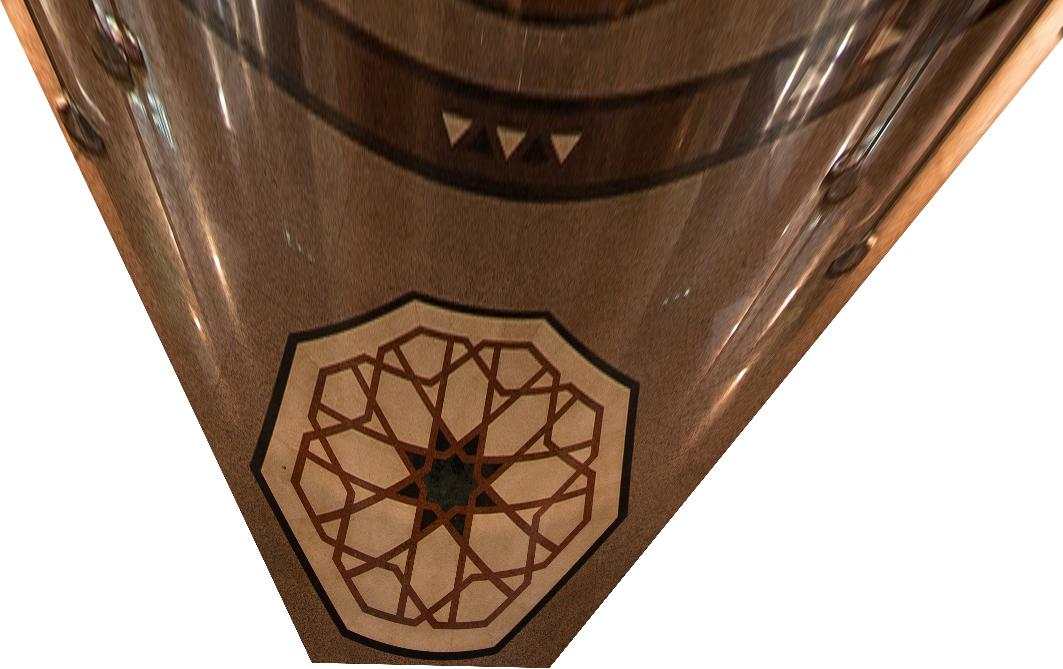}
\end{minipage}
\begin{minipage}{0.25\textwidth}
\centering
\includegraphics[width=\textwidth]{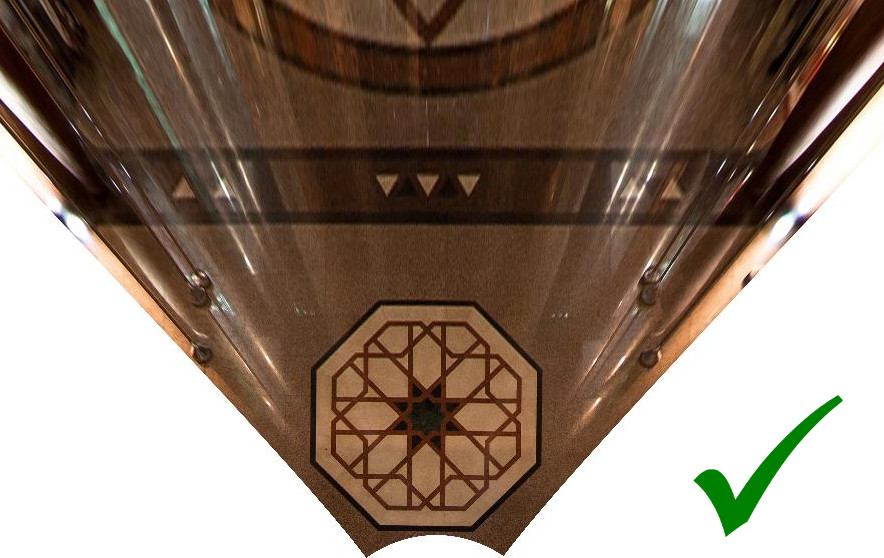}
\end{minipage}
\caption{\emph{Solver Comparison.} The state-of-the art solvers
$\mH_{22}\ve[l]\ve[u]\ve[v]\lambda$ and $\mH_{22}\vl
s_i$ \cite{Chum-ACCV10,Pritts-CVPR18} are compared with the proposed
solvers $\mH_{222}\vl \lambda$ and $\mH_{22}\vl$ on images containing
either translated or rigidly-transformed coplanar repeated patterns
with increasing amounts of lens distortion. (top) small distortion,
rigidly-transformed; (middle) medium distortion, translated; (bottom)
large distortion, rigidly-transformed. Accurate rectifications for all
images is only given by the proposed $\mH_{222}\vl \lambda$.} 
\label{fig:solver_comparison}
\end{figure*}

\clearpage

\bibliographystyle{splncs}

\appendix
\numberwithin{figure}{section}
\newpage
\addtocontents{toc}{\protect\setcounter{tocdepth}{2}}
\pagestyle{headings}

\setcounter{figure}{0}
\setcounter{table}{0}
\counterwithin{figure}{section}
\counterwithin{table}{section}

\title{Rectification from Radially-Distorted Scales Supplemental Material} 
\titlerunning{Supplemental}
\authorrunning{J. Pritts \etal}

\author{James Pritts\inst{1,2} \and Zuzana Kukelova\inst{2} \and Viktor Larsson\inst{3}\textsuperscript{,*} \and Ond{\v r}ej Chum\inst{2}}

\institute{Czech Institute of Informatics, Robotics and Cybernetics (CIIRC), CTU in Prague \and Visual Recognition Group (VRG), FEE, CTU in Prague \and Department of Computer Science, ETH Zürich, Switzerland}

\maketitle


\renewcommand{\thefootnote}{\fnsymbol{footnote}}
\footnotetext[1]{This work was done while Viktor Larsson was at Lund University.}
\renewcommand{\thefootnote}{\arabic{footnote}}

\section{Extended Experiments} 
\label{sec:extended_experiments}
\begin{figure*}[h]
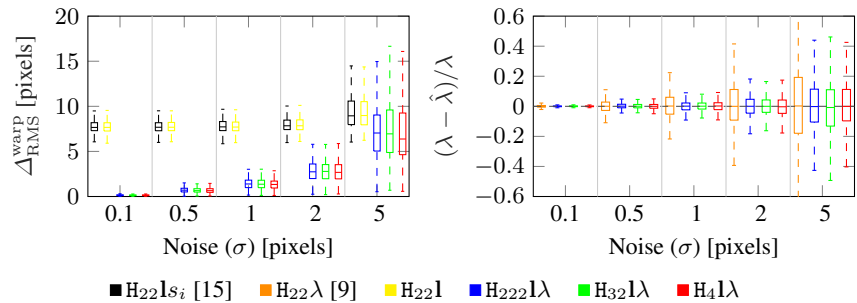

\raggedright
 \setlength\fwidth{0.35\columnwidth} \input{fig/ransac_rewarp_sensitivity_rt.tikz}
 \setlength\fwidth{0.35\columnwidth} \input{fig/ransac_rel_lambda_sensitivity_rt.tikz}
%
\centering \definecolor{mycyan}{rgb}{0,1,1}
\definecolor{myorange}{rgb}{1, 0.5490, 0}
\begin{tikzpicture}
\begin{customlegend}
[legend columns=-1,
legend style={draw=none,/tikz/every even column/.append style={column sep=0.3cm},cells={align=left}},
legend entries={$\mH_{22}\vl s_i$ \cite{Pritts-CVPR18}, $\mH_{22}\lambda$ [9], $\mH_{22}\vl$, $\mH_{222}\vl \lambda$,$\mH_{32}\vl \lambda$,$\mH_4 \vl \lambda$}]
    \addlegendimage{black,fill=black,only marks,mark=square*}            
    \addlegendimage{myorange,fill=myorange,only marks,mark=square*}     
    \addlegendimage{yellow,fill=yellow,only marks,mark=square*}                
    \addlegendimage{blue,fill=blue,only marks,mark=square*}          
    \addlegendimage{green,fill=green,only marks,mark=square*}
    \addlegendimage{red,fill=red,only marks,mark=square*}            
\end{customlegend}
\end{tikzpicture}  
\caption{\emph{Sensitivity Benchmark for Rigidly Transformed Coplanar Repeats.} 
Comparison of two error measures after 25 iterations of a
  simple \RANSAC for different solvers with increasing levels of white
  noise added to affine-frame correspondences that are rigidly
  transformed on the scene plane. (left) Reports the warp error as
  $\Delta_{\mathrm{RMS}}^{\mathrm{warp}}$ and (right) Reports the
  relative error of the estimated division-model parameter. The
  proposed undistorting
  solvers---$\mH_{222}\vl\lambda,\mH_{32}\vl\lambda,\mH_4\vl\lambda$---
  perform the best and exhibit similar noise characteristics with
  rigidly transformed coplanar repeats as with translated coplanar
  repeats (see \Fig\ref{fig:ransac_sensitivity_study}
  in \Sec\ref{sec:noise_sensitivity}).}
\label{fig:ransac_sensitivity_study_rt}
\end{figure*}

The extended experiments include the noise sensitivity experiments for
coplanar repeats that are rigidly transformed in the scene plane (see
\Fig\ref{fig:ransac_sensitivity_study_rt}). In \Sec\ref{sec:noise_sensitivity}
the sensitivity study was performed with conjugate translations so that
the state-of-the-art solvers of \cite{Pritts-CVPR18} could be
included. The sensitivity study for rigid transforms shown in
\Fig\ref{fig:ransac_sensitivity_study_rt} confirms
that the noise characteristics of the proposed solvers for
rigidly-transformed coplanar repeats are consistent with the results
shown in in \Fig\ref{fig:ransac_sensitivity_study} for conjugate
translations.

\begin{figure*}
\newcommand{\trirow}[1]{%
\setlength{\tabcolsep}{0.2cm}
\begin{tabular}{c}
\includegraphics[height=3.6cm]{suppimg/#1.jpg}\\
\includegraphics[height=3.6cm]{suppimg/#1_ud.jpg}\\
\includegraphics[height=2.2 cm]{suppimg/#1_rect_cropped.jpg}\\
\end{tabular}
}

\trirow{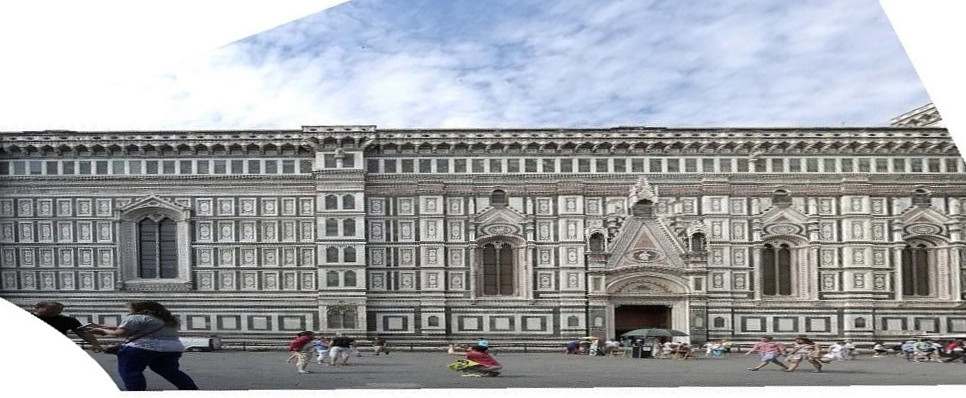}
\trirow{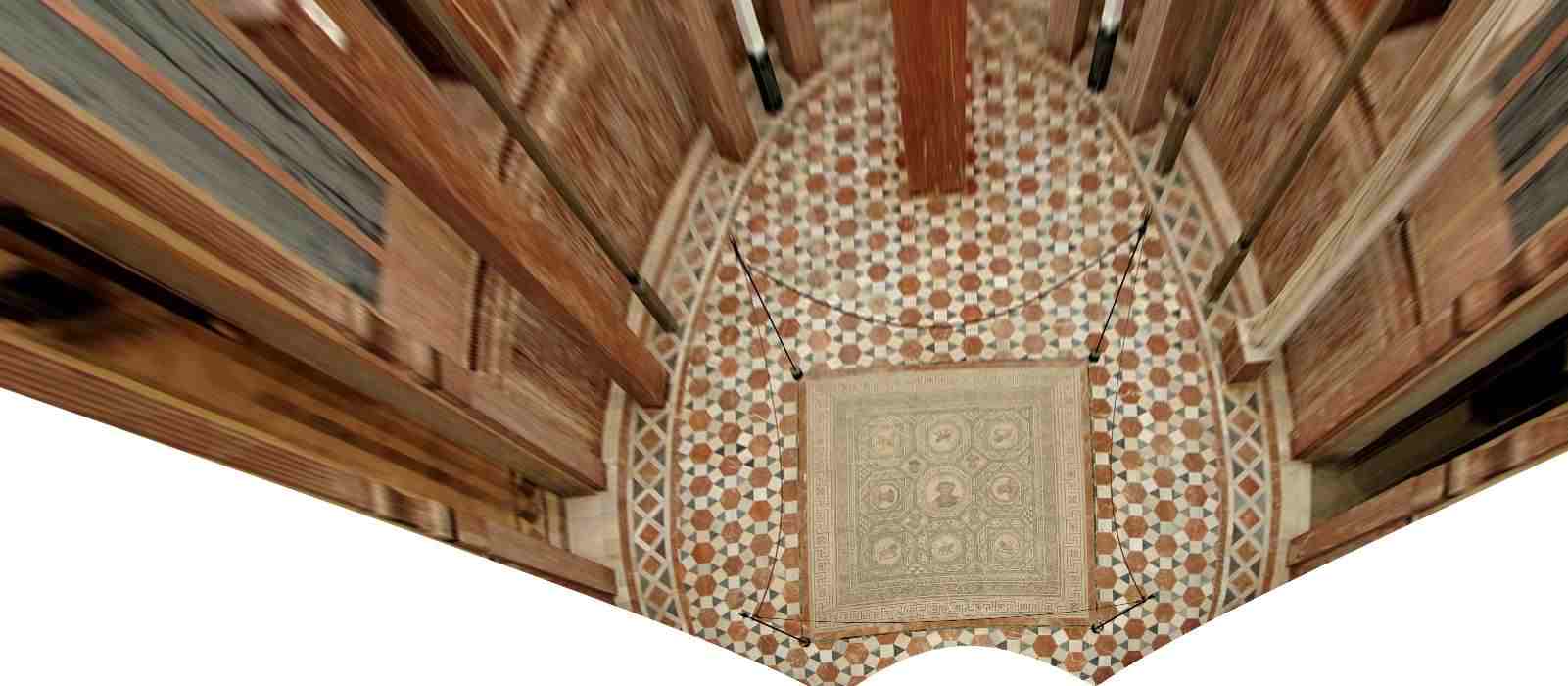}

\caption{ \emph{Fisheye Lenses.} The proposed method is tested
  on imagery taken 8mm and 10mm fisheye lenses. The division model
  used by \cite{Fitzgibbon-CVPR01} for radial lens distortion has only
  1 parameter, which limits its use for modeling extreme lens
  distortion. Even so, the proposed method gives reasonable solutions
  for affine rectification and undistortion on fisheye
  lenses. Rectification quality is also dependent on the coverage of
  the features extracted.  (top row) Input images (middle row)
  Undistorted images using $\mH_{222}\vl\lambda$ + LO (bottom row)
  Undistorted and rectified results.}
\label{fig:supp_fisheye_lenses}
\end{figure*}

\begin{figure*}
\newcommand{\trirow}[1]{%
\setlength{\tabcolsep}{0.2cm}
\begin{tabular}{c}
\includegraphics[height=3.9cm]{suppimg/#1.jpg}\\
\includegraphics[height=3.9cm]{suppimg/#1_ud.jpg}\\
\includegraphics[height=1.9 cm]{suppimg/#1_rect_cropped.jpg}\\
\end{tabular}
}

\trirow{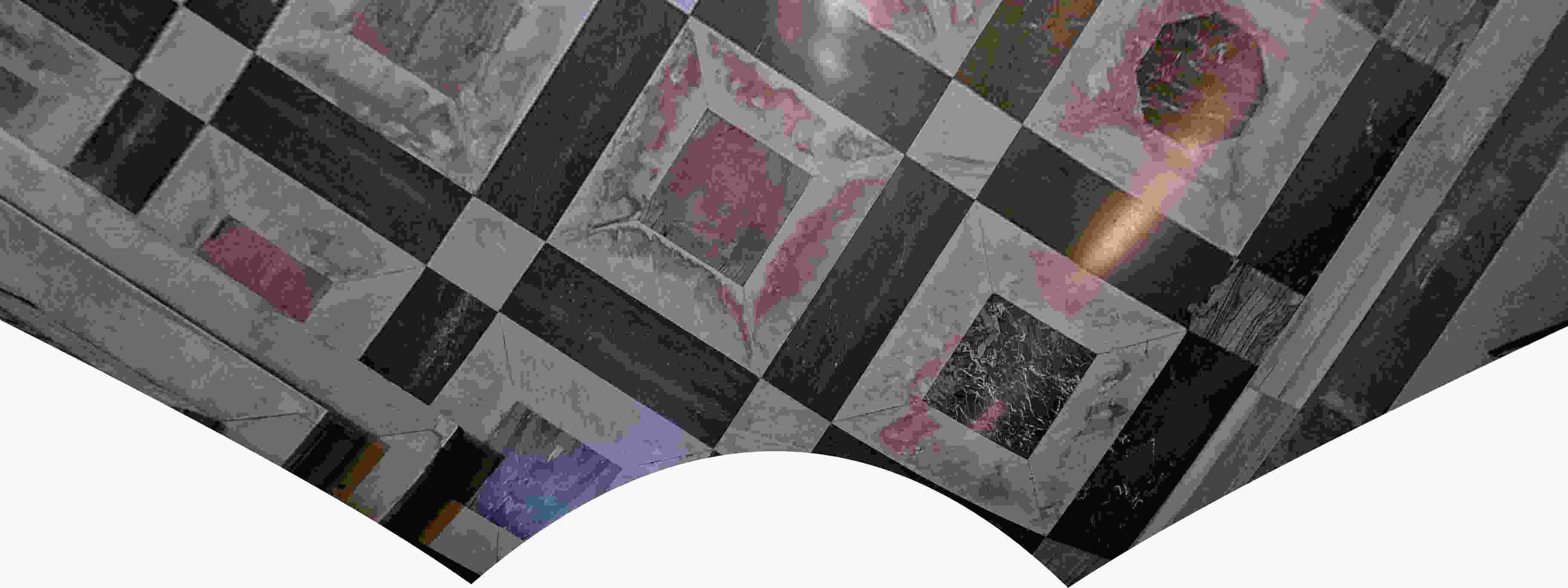}
\trirow{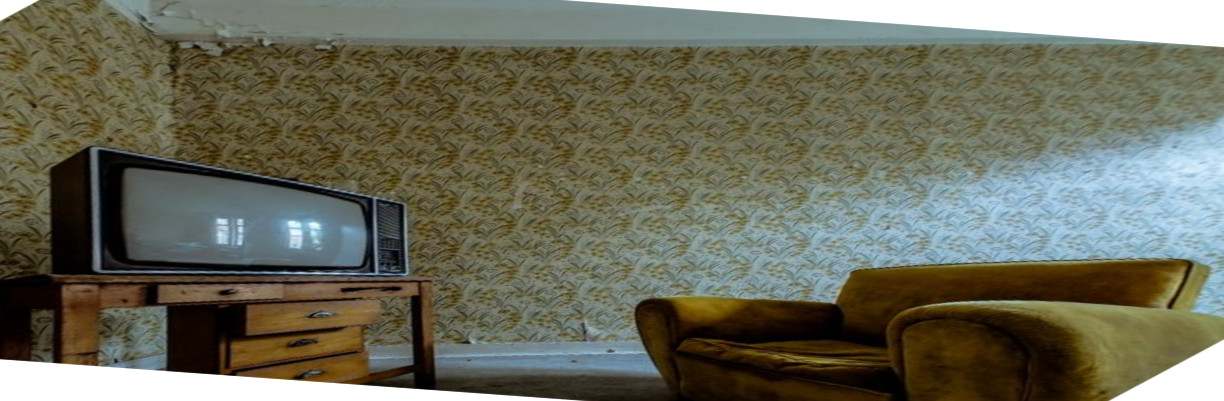}
\caption{\emph{More Fisheye Lenses.} The proposed method is tested
  on imagery taken from a 7.5mm fisheye lens. Some radial distortion
  is still visible in the undistorted image in the left column,
  perhaps due to poor coverage of affine-covariant regions across the
  image. (top row) Input images (middle row) Undistorted images using
  $\mH_{222}\vl\lambda$ + LO (bottom row) Undistorted and rectified
  results.}
\label{fig:supp_more_fisheye_lenses}
\end{figure*}

\begin{figure*}
\newcommand{\trirow}[1]{%
\setlength{\tabcolsep}{0.2cm}
\begin{tabular}{c}
\includegraphics[height=3.9cm]{suppimg/#1.jpg}\\
\includegraphics[height=3.9cm]{suppimg/#1_ud.jpg}\\
\includegraphics[height=4.1 cm]{suppimg/#1_rect.jpg}\\
\end{tabular}
}

\trirow{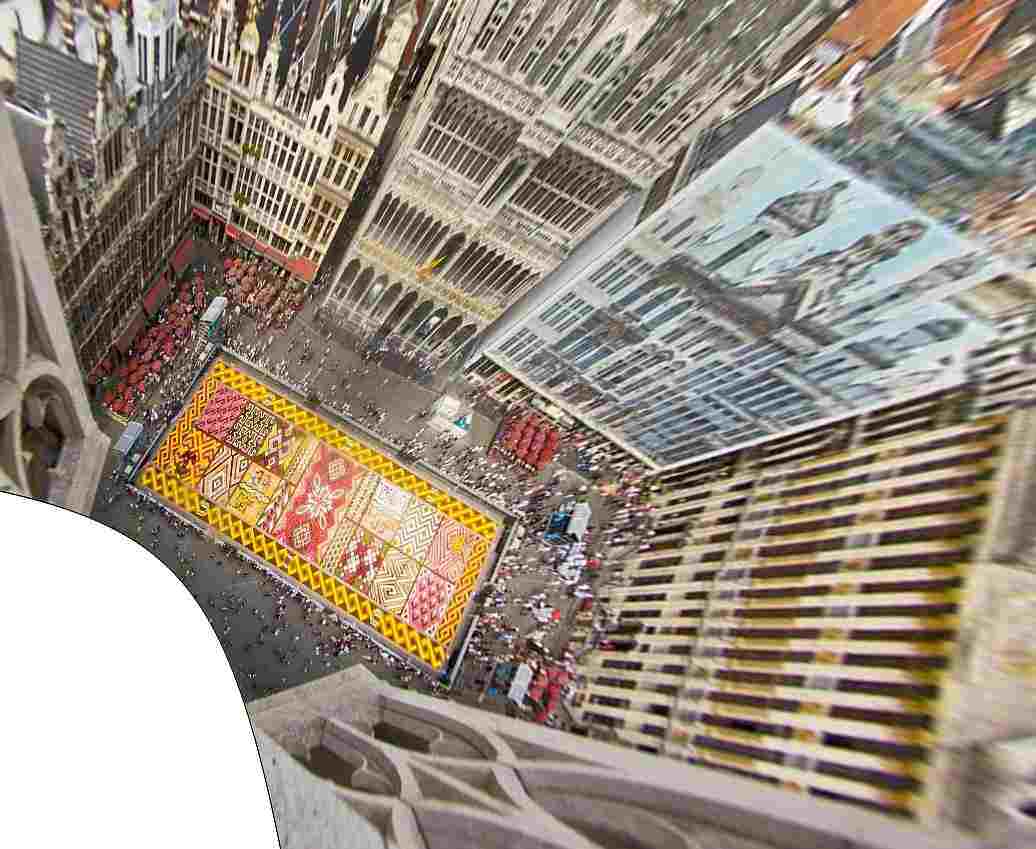}
\trirow{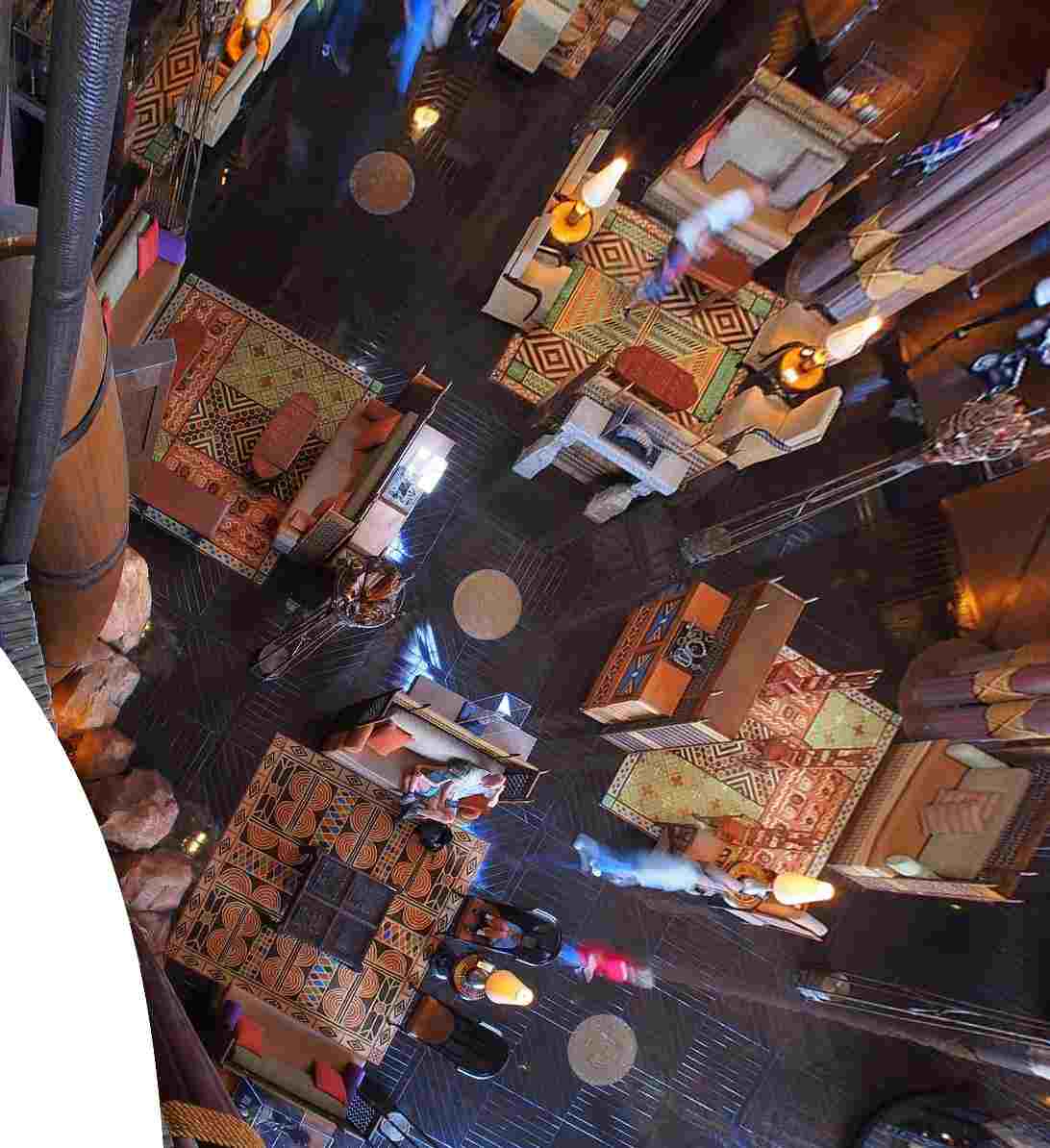}
\caption{\emph{Even More Fisheye Lenses.} Some radial distortion
  is still visible in the undistorted image in the left column,
  perhaps due to the fact that the estimation was from measurements
  from the texture near the center of distortion.  (top row) Input
  images: (left) 8mm lens, (right) 12mm lens. (middle row) Undistorted
  images using $\mH_{222}\vl\lambda$ + LO (bottom row) Undistorted and
  rectified results.}
\label{fig:supp_even_more_fisheye_lenses}  
\end{figure*}

Also included
in \Figs\ref{fig:supp_fisheye_lenses},\ref{fig:supp_more_fisheye_lenses},
and \ref{fig:supp_even_more_fisheye_lenses} are the undistorted and
rectified results for several images taken with fisheye lenses, which
further demonstrates the proposed method's effectiveness on diverse
and challenging image content. These images test the limits of the
one-parameter division model for modeling the extreme radial lens
distortion of fisheye lenses. Even so, the results are reasonable and
could be used to regress an initial guess at a higher parameter
fisheye model---\eg \cite{Kannala-PAMI06}---for use an input to a
non-linear refinement.

\end{document}